\PassOptionsToPackage{prologue,table}{xcolor}
\documentclass[sigconf,natbib=true,anonymous=false]{acmart}
\renewcommand\footnotetextcopyrightpermission[1]{} % removes footnote with conference information in first column
\usepackage{algorithm}
\usepackage{algpseudocode}
\algrenewcommand\textproc{\text}
\usepackage{makecell}
\usepackage{booktabs}
\usepackage{pifont}
\newcommand{\xmark}{\ding{55}}
\newcommand{\cmark}{\ding{51}}
\usepackage{multirow}
\usepackage{enumitem}
\usepackage{balance}
\usepackage{threeparttable}
\usepackage{amsmath,amsfonts,mathtools} % 已经包含了 amssymb 的大部分功能
\usepackage{colortbl}
\usepackage{mathrsfs}
\usepackage{float}
\usepackage{graphicx}
\usepackage{subfigure}
\usepackage{hyperref}
\usepackage{tabularx, booktabs}
\usepackage{tikz}
\usepackage{arydshln}
\usepackage{CJKutf8}
\newcommand{\bluefont}[1]{ {\color{blue}{#1}}}
\newcommand{\redfont}[1]{{\textcolor{red}{#1}}}
\newcommand{\greenfont}[1]{{\textcolor{green}{#1}}}
\AtBeginDocument{%
  \providecommand\BibTeX{{%
    \normalfont B\kern-0.5em{\scshape i\kern-0.25em b}\kern-0.8em\TeX}}}

\usepackage{tikz}
\usepackage{setspace}
\usepackage[most]{tcolorbox}
\definecolor{uc_color}{rgb}{0.99,0.24,0.63}
\definecolor{hc_color}{rgb}{0.02,0.51,0.51}
\definecolor{tc_color}{rgb}{0.99,0.55,0.09}

\tikzstyle{mybox} = [draw=black, very thick,
    rectangle, rounded corners, inner sep=10pt, inner ysep=13pt]
\tikzstyle{fancytitle} =[fill=black, text=white]

\begin{document}
\begin{CJK*}{UTF8}{gkai}
\title{HyKGE: A Hypothesis Knowledge Graph Enhanced Framework for Accurate and Reliable Medical LLMs Responses}

\author{Xinke Jiang\textsuperscript{*}, Ruizhe Zhang\textsuperscript{*}, Yongxin Xu}\authornote{Xinke Jiang, Ruizhe Zhang, and Yongxin Xu contributed equally to this research. {\{xinkejiang, ruizezhang, xuyx}\} @stu.pku.edu.cn }
\author{Rihong Qiu, Yue Fang, Zhiyuan Wang, Jinyi Tang, Hongxin Ding}
\author{Xu Chu\textsuperscript{\textdagger}, Junfeng Zhao\textsuperscript{\textdagger}\textsuperscript{\textdaggerdbl}, Yasha Wang}
\affiliation{
  \institution{Key Laboratory of High Confidence Software Technologies (Peking University) \\ Ministry of Education; School of Computer Science, Peking University}
  \city{Beijing}
  \country{China}
}
\authornote{Corresponding authors.}
\authornote{Junfeng Zhao is also at the Big Data Technology Research Center, Nanhu Laboratory, 314002, Jiaxing.}
\email{}

\renewcommand{\shortauthors}{Jiang, Zhang and Xu et al.}

\begin{abstract}
% 从检索前和检索后说
In this paper, we investigate the retrieval-augmented generation (RAG) based on Knowledge Graphs (KGs) to improve the accuracy and reliability of Large Language Models (LLMs). 
Recent approaches suffer from insufficient and repetitive knowledge retrieval, tedious and time-consuming query parsing, and monotonous knowledge utilization.
To this end, we develop a \underline{\textbf{Hy}}pothesis \underline{\textbf{K}}nowledge \underline{\textbf{G}}raph \underline{\textbf{E}}nhanced (\textbf{HyKGE}) framework, which leverages LLMs' powerful reasoning capacity to compensate for the incompleteness of user queries, optimizes the interaction process with LLMs, and provides diverse retrieved knowledge.
Specifically, HyKGE explores the zero-shot capability and the rich knowledge of LLMs with Hypothesis Outputs to extend
feasible exploration directions in the KGs, as well as the carefully curated prompt to enhance the density and efficiency of LLMs' responses. 
Furthermore, we introduce the
HO Fragment Granularity-aware
Rerank Module to filter out noise while ensuring the balance between diversity and relevance in retrieved knowledge. 
Experiments on two Chinese medical multiple-choice question datasets and one Chinese open-domain medical Q\&A dataset with two LLM turbos demonstrate the superiority of HyKGE in terms of accuracy and explainability.
\end{abstract}




\maketitle

\balance
% \vspace{-0.7cm}
\section{Introduction}
\label{intro}
\textbf{Large Language Models (LLMs)}, such as ChatGPT~\cite{ChatGPT} and GPT-4~\cite{OpenAI2023GPT4TR}, have achieved remarkable progress in pivotal areas. By undergoing pre-training on massive text corpora and aligning fine-tuning to follow human instructions~\cite{ziegler2020finetuning,wang2023selfinstruct}, they have recently demonstrated exceptional performance in a range of downstream tasks~\cite{kaplan2020scaling}. These achievements underscore the vast potential of LLMs in understanding and generating natural language~\cite{vu2024gptvoicetasker}, especially in the medical domain~\cite{Kraljevic_Bean_Shek_Bendayan_Hemingway_Yeung_Deng_Baston_Ross_Idowu_et,Yang_Zhao_Zhu_Zhou_Xu_Jia_Zan_2023,Zhu_Togo_Ogawa_Haseyama,Wang_Duan_Lam_Xu_Chen_Liu_Pang_Tan_2023,Xiong_Wang_Zhu_Zhao_Liu_Huang_Wang_Shen,Wang_Liu_Xi_Qiang_Zhao_Qin_Liu_2023,Zhang_Chen_Jiang_Yu_Chen_Li_Chen_Wu_Zhang_Xiao_et,Pal_Sankarasubbu,DISCMedLLM}.
Despite the advancements of fine-tuning, they still encounter significant challenges, including the difficulty in avoiding factual inaccuracies (i.e., hallucinations and limited explainability)~\cite{ji2023survey,cao-etal-2020-factual,10.1145/3571730}, data constraints (i.e. token resource limit, high training costs, and privacy concerns)\footnote{https://www.youtube.com/watch?v=ahnGLM-RC1Y}, catastrophic forgetting~\cite{GqAO}, outdated knowledge~\cite{he2022rethinking}, and a lack of expertise in handling specific domains or highly specialized queries~\cite{kandpal2023large}.
% According to an OpenAI report\footnote{https://www.youtube.com/watch?v=ahnGLM-RC1Y} and ~\cite{zhao2023survey,taylor2022galactica}, which identify several key challenges in maximizing the potential of LLMs , including \textit{hallucinations, privacy concerns, high training costs, limited explainability, token resource constraints, and difficulties in assimilating new and domain-specific knowledge}.
This undermines their reliability in areas where accountability and trustworthiness are crucial and infallible in the medical area~\cite{ji2023survey,song2024typing,LAI2023104392}.

\textbf{Retrieval-Augmented Generation (RAG)}, enhances content generation by retrieving external information, reduces factual errors in knowledge-intensive tasks with the help of external knowledge and is seen as a promising solution to address incorrect answers, hallucinations, and insufficient interpretability~\cite{izacard2022atlas,asai2023self,asai2023retrieval}.
Among the numerous external information sources~\cite{yang2024faima}, \textbf{knowledge graphs} (KGs), as a structured data source refined and extracted through advanced information extraction algorithms, can provide higher quality context. Compared to documents, KGs embody structured knowledge~\cite{zhong2023comprehensive,Ji_2022}, providing succinct content and facilitating the analysis of intricate relationships among entities, leading to advanced inference capabilities and enabling extrapolation for efficient knowledge retrieval. They are considered by many research works to improve the accuracy and reliability of answers provided by LLMs~\cite{kg_review,wen2023mindmap}. 
% However, due to the inherent characteristics of query and KG, the performance of RAG on KG is always not satisfactory~\cite{kgrag_arxiv}.
% Thus, how to properly retrieve knowledge (\textbf{pre-retrieval phase}) and how to handle the retrieved knowledge (\textbf{post-retrieval phase}) still remain huge challenges: 
However, the gap between unstructured user queries of inconsistent quality and structured, high-quality KGs~\cite{kgrag_arxiv} poses significant challenges on how to properly parse user intent for improving the robustness of retrieved knowledge (\textbf{pre-retrieval phase}) and how to handle the abundant retrieved knowledge (\textbf{post-retrieval phase}), which are detailed as follows:

\textbf{Challenge I: } \textbf{At the pre-retrieval phase, previous works suffer from how to 
parse user intent and retrieve reasonable knowledge based on varying-quality user query.} % 瑞哲看一下这里
Some works are based on the Retrieve-Read framework, which initially obtains knowledge through dense vector retrieval according to user queries~\cite{ma2023chainofskills,qu2021rocketqa,yu2022kgfid}. However, they are stricken with issues such as unclear expressions and lack of semantic information in the user's original query. 
This misalignment between the semantic spaces of user queries and high-quality structured knowledge leads to the retrieval of knowledge that is of insufficient quality and may contain redundant information and noise~\cite{barnett2024seven}. 
In addition, the excessive redundant knowledge can lead to a waste of token resources, and the response speed of LLMs will drop sharply, which adversely damages the performance in real-world applications~\cite{finardi2024chronicles}.
\begin{figure*}[t]
  \centering
  \vspace{-0.33cm}
\includegraphics[width=\textwidth]{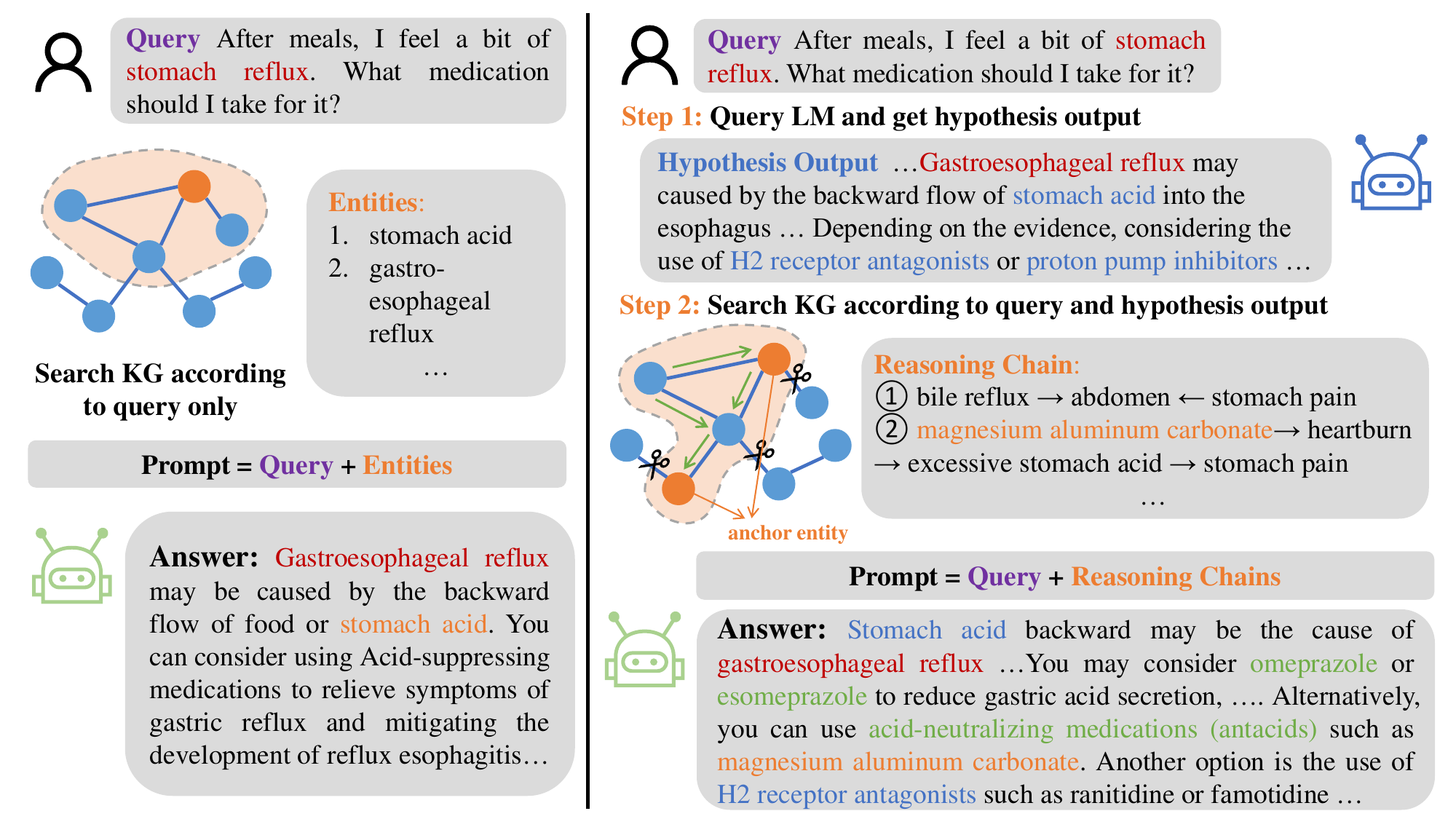}
  \vspace{-0.2cm}
  \caption{{(a) KGRAG (Left)}. Basic KGRAG extracts key entities from user queries and searches for corresponding entities within KG, which are then fed into LLMs along with the query. {(b) HyKGE (Right)}. HyKGE first queries LLMs to obtain hypothesis output and extracts entities from both the hypothesis output and the query. Then HyKGE retrieves reasoning chains between any two anchor entities and feeds the reasoning chains together with the query into LLMs. }
  \vspace{-0.25cm}
  \label{fig:intro.png}
\end{figure*}

\textbf{Challenge II: } \textbf{At the pre-retrieval phase, how to align user intent with high-quality structured knowledge while reducing interactions with LLMs remains an unresolved issue.} Some works enable LLMs to step-by-step utilize knowledge to enhance intent parsing and inference of user queries. They facilitate the acquisition of planning and reflective abilities in LLMs' interactions with KGs through multi-round chain-of-thought requests~\cite{sun2023thinkongraph,li2023chainofknowledge,wang2023knowledgedriven,asai2023self,yu2023chainofnote}. However, they are constrained by the expensive time overhead of multiple interactions with LLMs and the cumulative errors in the distributed reasoning process.
% In summary, 

\textbf{Challenge III: }
\label{c 3}\textbf{At the post-retrieval phase, previous studies often struggle with the dilemma of balancing the diversity and relevance of the retrieved knowledge.} Recent post-retrieval models typically apply similarity filtering or a reranking approach in response to user queries to prune retrieved results~\cite{toro2023dynamic,finardi2024chronicles,cuconasu2024power}. However, user queries often exhibit notably monotonous properties and sparsely distributed keywords because the prevalence of natural language descriptions will tend to dilute its concentration~\cite{Breuer_2023}. Conversely, KGs are characterized by their inherently structured nature, resulting in a high knowledge density within retrieved results. As a consequence, pruning knowledge solely based on the user query can lead to a misalignment in knowledge density and the final result is often highly correlated yet excessively repetitive, significantly diminishing the efficacy of RAG. Therefore, one of the primary challenges in the post-retrieval phase is to balance the trade-off between relevant knowledge and diverse ones~\cite{Breuer_2023}. 

To cope with these challenges, we put forward the \textbf{\underline{Hy}}pothesis \textbf{\underline{K}}nowledge \textbf{\underline{G}}raph \textbf{\underline{E}}nhanced (\textbf{HyKGE}) framework, a novel method based on the hypothesis output module (\texttt{HOM}) ~\cite{hyde} to explore, locate, and prune search directions for accurate and reliable LLMs responses in pre-retrieval phase and greatly preserve the relevance and diversity of search results at in post-retrieval phase.
\underline{\textbf{i)}} Specifically, in the \textbf{pre-retrieval phase}, our key idea is that the zero-shot capability and rich knowledge of LLMs can compensate for the incompleteness of user queries, facilitating alignment with high-quality external knowledge.
For example, when facing the question ``\textit{After meals, I feel a bit of stomach reflux. What medicine should I take? }'', if retrieval is based solely on the key entity ``\textit{stomach reflux }'' as illustrated in Figure \ref{fig:intro.png}(a), a large amount of noise will be introduced due to the broad semantics of the entity.
However, if LLMs are guided to explore how to solve the problem, they will provide additional clues related to ``\textit{stomach acid }'',  ``\textit{H2 receptor antagonists }'' and ``\textit{proton pump inhibitors }'' as illustrated in Figure \ref{fig:intro.png}(b), based on the knowledge acquired during their pre-training and instruction fine-tuning phases, offering exploration directions for retrieval on the KGs. 
\underline{\textbf{ii)}} Meanwhile, HyKGE utilizes the flexibility of natural language in prompts to set constraints, enabling LLMs to provide as comprehensive information as possible when outputting hypothesis results, thereby reducing the number of interactions and improving efficiency. \underline{\textbf{iii)}} In the \textbf{post-retrieval stage}, to further enhance the alignment between user queries and external knowledge inference paths, we propose a Hypothesis Output-based (HO) Fragment Granularity-aware, which utilizes multiple short snippets from the hypothesis outputs as well as the user query to rerank and filter the retrieved knowledge, greatly avoiding the filtering of diverse knowledge. It ensures fine-grained interaction and filtering while addressing the issue of imprecise matching between monotonous and sparse text (user query) with multi-element and dense text (retrieved knowledge). Through comprehensive experiments, our main contributions can be summarized as follows:
\begin{itemize} [leftmargin=*]
    \item  At the pre-retrieval phase, 
    we leverage the zero-shot capability of LLMs to obtain an exploratory and hypothesis output, transforming the incomplete and non-professional nature of user queries. Corresponding anchor entities are then identified from the hypothesis output on the KGs, providing a direction for exploration and pruning retrieval space. Simultaneously, we utilize the knowledge chains to rectify errors and illogicalities in the hypothesis outputs, mitigating hallucinations and false knowledge problems. 
    \item At the post-retrieval stage, we propose a HO Fragment Granularity-aware rerank module to further enhance the knowledge density alignment between the retrieved reasoning chains and hypothesis outputs at a finer granularity, greatly preserving relevant yet diverse knowledge through the idea of divide-and-conquer.
    % rearrange and integrate the retrieved knowledge , further enhancing the alignment between user queries and external knowledge reasoning chains through the idea of divide-and-conquer.
    \item We validate the superiority of the HyKGE through various observations by experiments on two Chinese medical multiple-choice question datasets and one Chinese medical open-domain Q\&A dataset with two LLM turbos. This integration of LLMs and KGs addresses key challenges in medical LLMs, notably in accuracy and explainability, and has potential applications in improving medical consultation quality, diagnosis accuracy, and expediting medical research. 
\end{itemize}

\section{Related Work}
% 3. RAG的应用和进展：探讨RAG技术在其他研究中的应用，特别是在语言模型中的应用。主要讨论KG-RAG和文档RAG
\paragraph{\textbf{Retrieval-Augmented Generation. }} 
RAG incorporates the external knowledge retrieval component via prompt engineering to achieve more factual consistency, enhancing the reliability and interpretability of LLMs' responses~\cite{lewis2021retrievalaugmented}. 
% These approaches leverage a retriever model to identify a set of documents that are relevant to user queries from extensive knowledge corpora (i.e.,  Wikipedia or document repositories). 
% Subsequently, a reranker model extracts the most relevant contents from the retrieved documents, which is then fed into the reader model (i.e. LLMs) for further processing~\cite{kg_review}. 
Classic RAG methods leverage retriever models to source relevant documents from large knowledge corpora~\cite{xu2024retrieval}, followed by reranker models that distill contents and reader models for further processing~\cite{kg_review,sarthi2024raptor}. Despite advancements in retriever~\cite{qu2021rocketqa,ma2023chainofskills,chen2024improving} and reranker efficiency~\cite{cheng2021unitedqa,yu2022kgfid}, they still encounter difficulty in acquiring high-quality datasets for training query-document pair retrievers or limited information in user queries which weakens their generalization capability~\cite{gao2022precise}. 
Moreover, some researches focus on fine-tuning reader LLMs, applying instruction-tuning with retrieved knowledge or RAG API calls~\cite{luo2023sail,izacard2022atlas,asai2023selfrag,yoran2023making,wang2024rear,zhang2023knowledgeable,lin2023radit}. However, creating such datasets is also challenging due to the need for manual label correction, which in turn, may erode LLMs generalization capabilities and cause catastrophic forgetting in routine Q\&A tasks.
% Although fine-tuning the LLMs through instructions can promote LLMs' better understanding of the knowledge of RAG or clarify when performing RAG, constructing such data poses challenges, as it requires humans to manually correct the supervised label contents which are not naturally obtainable. 

Beyond optimizing submodels, HyDE~\cite{gao2022precise} introduces 
% an approach leveraging hypothesis outputs generated by instruction-following LLMs based on user query. 
% These hypothesis outputs are then used as input for the retriever, which effectively enhances retrieval performance in zero-shot scenarios.
an innovative method where instruction-following LLMs generate hypothesis documents based on user queries to enhance retriever performance, particularly in zero-shot scenarios. 
Other methods like CoN~\cite{con} and CoK~\cite{cok} involve LLMs in note-making and step-wise reasoning verification through customized prompts, and greatly rely on frequent interactions with LLMs.
However, such an approach is excessively inefficient for deployment in real-world Q\&A scenarios.

Our HyKGE, uses LLM hypothesis output for exploratory directions in KGs and corrects model errors using graph reasoning chains during pre-retrieval, and applies fine-grained alignment in post-retrieval to maintain effective, diverse knowledge, enhancing retrieval efficiently without fine-tuning or excessive interactions.

\paragraph{\textbf{Knowledge Graph Query-Answer. }}Compared to knowledge stored in document repositories~\cite{izacard2022unsupervised}, the knowledge contained within KGs has the advantages of being structured and inferable, rendering it a more suitable source for supplementing LLMs~\cite{luo2023reasoning,jiang2023reasoninglm,liu2021kgbart,kang2023knowledge,Sen2023KnowledgeGL,Varshney2023KnowledgeGM}. However, how to design a retriever to extract knowledge from KGs and how to design interaction strategies between LLM and KGs are still in the exploratory stage\footnote{https://siwei.io/talks/graph-rag-with-jerry/1}. 
% is the base approach for combining Knowledge Graph with LLMs leveraging Graph RAG, but it lacks the processing of redundant knowledge. 
KGRAG ~\cite{kgrag_arxiv} uses the user query as a reference for retrieval in KGs, which suffers from misalignment between high-quality structured knowledge and varying-quality queries. Semantic parsing methods allow LLMs to convert the question into a structural query (e.g., SPARQL), which can be executed by a query engine to derive the answers on KGs~\cite{sun2020sparqa,li2023fewshot,cok}. However, these methods depend heavily on the quality of generated query sentences, displaying subpar performance when confronted with intricate queries.

\section{Preliminaries}
% \subsection{Problem Formulation}
\begin{definition} [Knowledge Graph] Given a medical knowledge graph, denoted by $\mathcal{KG} = (\mathcal{E}, \mathcal{R}, \mathcal{T}, \mathcal{D}, \mathcal{N})$, where $\mathcal{E} = \{e_1, \ldots, e_N\}$ is the set of entities, $\mathcal{R} = \{r_1, \ldots, r_P\}$ is the set of relations, and $\mathcal{T} = \{(e_{t_i^{head}}, r_{t_i}, e_{t_i^{tail}}) \mid 1 \leq i \leq T, e_{t_i^{head}}, e_{t_i^{tail}} \in \mathcal{E}, r_{t_i} \in \mathcal{R}\}$ is the set of head-relation-tail triplets (facts). Additionally, ${d}_i \in \mathcal{D}$ represents the entity description of $e_i$, and $\mathcal{N}_v = \{(r, u) \mid (v, r, u) \in \mathcal{T}\}$ stands for the set of neighboring relations and entities of an entity $v$.
\end{definition}

% \begin{definition} [Token Vocabulary \& User Query] 
% Given a $\mathcal{V} = \{ [\text{MASK}], [\text{CLS}], [\text{EOS}], w_1, \ldots, w_V \}$ be the token vocabulary set, where $[\text{MASK}]$ is the mask token for masked language modeling, and $[\text{CLS}]$ and $[\text{EOS}]$ represent the beginning and end tokens of the sentence, respectively. The user query is represented as a sequence of tokens $\mathcal{Q} = (q_1, q_2, \ldots, q_M)$, where $q_j \in \mathcal{V}$ is the $j$-th token in the query, and $M$ is the maximum number of tokens.

% \end{definition}
\begin{definition} [Knowledge Graph Retrieval] 
Knowledge Graph Retrieval~\cite{9228848} is a module that focuses on efficiently retrieving relevant information from $\mathcal{KG}$ based on the user query \(\mathcal{Q}\). In KGs, information is represented as entities, relations, and attributes, forming a structured network. The goal of retrieval is to find entities or relationships that additionally supply knowledge for LLMs. Particularly, we retrieve knowledge from the matched entities $\{e_j\}$ such as entity names, entity types, descriptions $\{d_j\}$ and even triplets or subgraphs $\mathcal{G}_{e_j} = (e_j, \mathcal{T}_j, d_j)$.
% However, subgraphs have too much noise, so some researchers take connected subgraphs, entity-overlapping subgraphs, or paths between entities into consideration. 
\end{definition}

\section{Method}
\begin{figure*}[ht]
  \centering
  \vspace{-0.3cm}
\includegraphics[width=\textwidth]{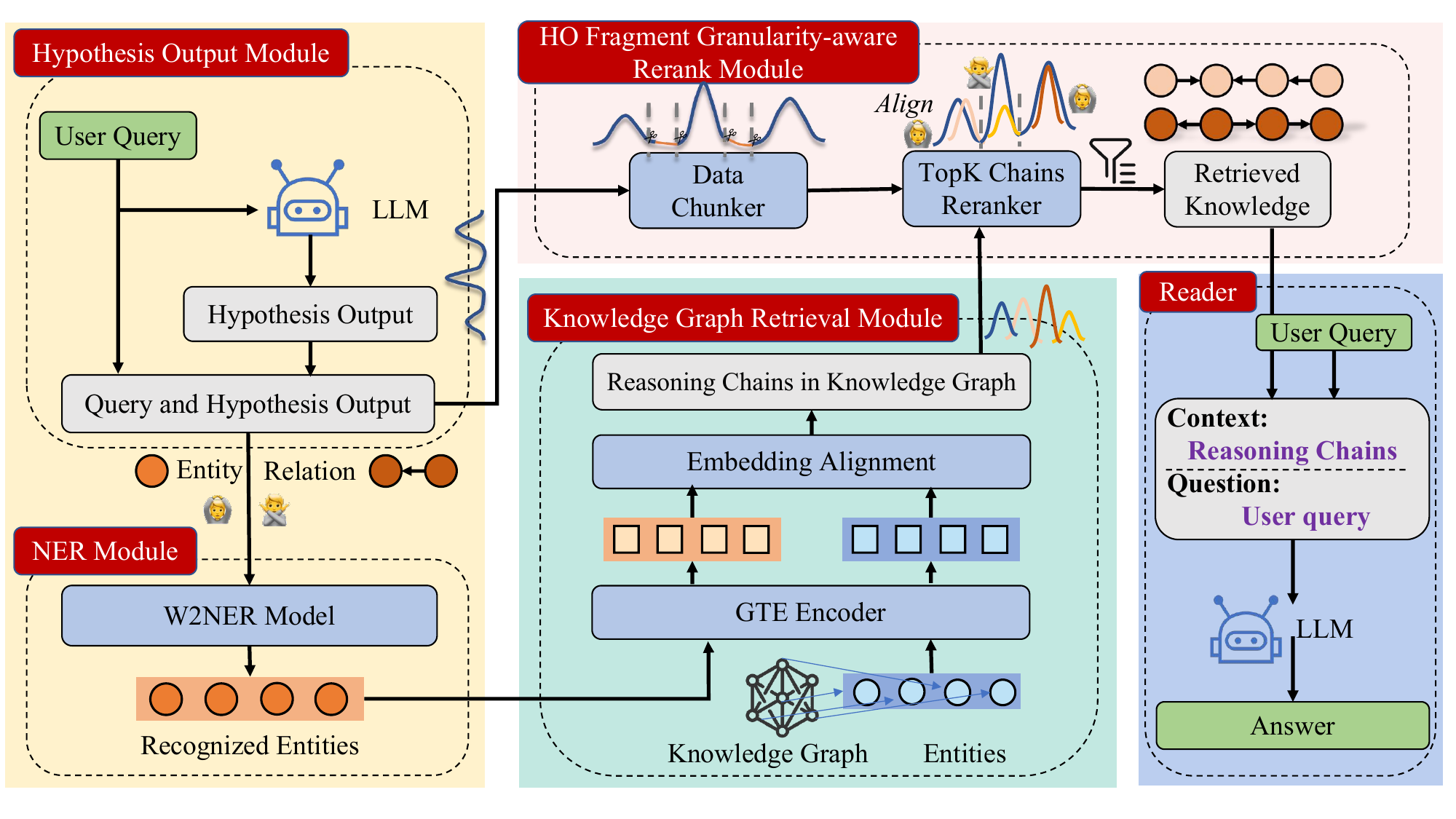}
\vspace{-0.35cm}
  \caption{The overall framework of HyKGE. HyKGE first feeds the user query ($\mathcal{Q}$) through the LLMs and obtains Hypothesis Output ($\mathcal{HO}$). Then through the NER Module, a W2NER model is applied to recognize entities and isolate relations. Through GTE Encoder, these recognized entities are then linked with entities in KGs. After that, HyKGE extracts three types of relevant reasoning chains from KGs.  
  % used by the Knowledge Graph Retrieval Module to extract relevant reasoning chains from KG, facilitated by a GTE Encoder and inner product calculations.
  Then, because of the sparseness of $\mathcal{Q}$, in the HO Fragment Granularity-aware Rerank Module, 
  HyKGE chunks $\mathcal{Q}$ and $\mathcal{HO}$ and align with reasoning chains via a TopK Chains Reranker, to eliminate irrelevant knowledge. Finally, we organize retrieved knowledge with the user query and obtain answers through LLM Reader.}
  \vspace{-0.23cm}
  \label{fig:method.png}
\end{figure*}
In this section, we detail our proposed HyKGE, and the overall framework is illustrated in Figure~\ref{fig:method.png}. In general, we will discuss our model from the four pipeline architectures:
\begin{itemize}[leftmargin=*,noitemsep,topsep=2pt]
    \item \textit{\textbf{Pre-Retrieval Phase}} includes the Hypothesis Output Module (\texttt{HOM}) and the NER Module (\texttt{NM}). \texttt{HOM} leverages LLMs to obtain hypothesis output by exploring possible answers. Then \texttt{NM} extracts medical entities from HO and the user query.
    \item \textit{\textbf{Retrieval on Knowledge Graph}} utilizes the extracted entities as anchors to search three distinct types of reasoning chains interlinking these anchors, providing relevant and logical knowledge.
    \item \textit{\textbf{Post-Retrieval Phase}} utilizes the HO Fragment Granularity-aware rerank approach. First, the hypothesis output and the user query are segmented into discrete fragments, and subsequently, we rerank the retrieved reasoning chains based on the fragments.
    \item \textit{\textbf{LLM Reader}} is fed with the user query and the pruned retrieved reasoning chains, organized with carefully designed prompts.
\end{itemize}
Next, we will delineate each phase in detail in the following subsections and state the overall process in Section~\ref{sec:porcess}. 

\subsection{Pre-Retrieval Phase}
% 讲更多关于HO的故事，为什么要用HO？为什么要NER，它俩是怎么配合的？这个故事要讲清楚: 传统的
Firstly, we let LLMs generate hypothesis outputs ($\mathcal{HO}$) in response to user query $\mathcal{Q}$, and then use the NER model to extract entities from both $\mathcal{HO}$ and $\mathcal{Q}$. During this process, LLMs utilize inherent medical knowledge to explore potential answers. Although $\mathcal{HO}$ may contain factual errors or hallucinations between entities, {the NER Module focuses solely on the extraction of entities while disregarding the relations,} thus significantly isolating the correlation among medical entities. The subsequent graph retrieval phase (c.f. Section \ref{sec: kg retrieval}) searches the correct reasoning chains to discern and reintegrate the relationships between medical entities, avoiding LLMs' shortages. The combination of \texttt{HOM} and \texttt{NM} provides us with a direction for exploration and identifies corresponding anchors in the KGs to guide subsequent graph retrieval, ensuring consistency and effectiveness in information processing.

% Firstly, the user's query initially undergoes processing through the primary LLM, yielding a hypothesis output. Throughout this procedure, LLM assimilates its intrinsic knowledge, thereby furnishing us with a trajectory for exploration. Subsequently, we employ a knowledge extraction model (NER) to extract medical entities from the hypothesis output.
\subsubsection{Hypothesis Output Module}
To enhance the quality of $\mathcal{HO}$, due to LLMs' robust reasoning abilities and potential as knowledge bases, we meticulously design instructions to guide LLMs in a step-by-step exploration and thoughtful consideration of problems, as illustrated in Figure \ref{tab:prompts} (Up.).
Here, the prompt (a textual instruction) is denoted as $\mathcal{P}_{\texttt{HO}}$, and $\mathcal{Q}$ to $\mathcal{HO}$ as:
\begin{equation}
\mathcal{HO} = \texttt{LLM}(\mathcal{Q}~|~\mathcal{P}_{\texttt{HO}}).
    \label{eq:ho}
\end{equation}
Thus, in light of the powerful reasoning abilities as well as the knowledgeable medical cognition, galore medical knowledge relevant to $\mathcal{Q}$ is discovered.
\begin{figure*}[ht!]
\caption{The prompt formats of (Up.) Hypothesis Output Module and (Down.) LLM Reader.}
\centering
\begin{tabular}{c} 
% Palmyra-Med
\begin{tikzpicture}
\node [mybox] (box){%
    \begin{minipage}{0.935\textwidth}
    \footnotesize
    \setstretch{1.2}
        \begin{flushleft} % Center the content
        {\textbf{\#\#\# Task Description:}} \\You are a medical expert. Please write a passage to answer \textbf{[User Query]} while adhering to \textbf{[Answer Requirements]}.
        
        \vspace{5pt}
        
        {\textbf{\#\#\# Answer Requirements:}} \\1) Please take time to think slowly, understand step by step, and answer questions. Do not skip key steps.  \\2) Fully analyze the problem through thinking and exploratory analysis.

           \vspace{5pt}
        
        \textbf{\#\#\# \{\{ User Query \}\}}
        \end{flushleft}
    \end{minipage}
};
\node[fancytitle, rounded corners, right=10pt] at (box.north west) {The Prompt Format of Hypothesis Output Module ($\mathcal{P}_{\texttt{HO}}$)};
\end{tikzpicture}%
\end{tabular}
\begin{tabular}{c} 
% Palmyra-Med
\begin{tikzpicture}
\node [mybox] (box){%
    \begin{minipage}{0.935\textwidth}
    \footnotesize
    \setstretch{1.2}
        \begin{flushleft} % Center the content
        {\textbf{\#\#\# Task Description:}} \\You are a medical expert. Based on relevant medical \textbf{[Background Knowledge]} and your medical knowledge, provide professional medical advice for \textbf{[User Query]} while adhering to \textbf{[Answer Requirements]}.
        
        \vspace{5pt}
        
        {\textbf{\#\#\# Answer Requirements:}}\\ 1) Take time to think slowly, understand step by step, and answer questions. \\2) Clearly state key information in the answer and provide direct and specific answers to user questions.
        
        \vspace{5pt}
        
        \textbf{\#\#\# \{\{ Background Knowledge \}\}} 
        \\ The retrieved knowledge chains are: 
        \tcbox[on line, boxsep=1pt, left=0pt,right=0pt,top=0pt,bottom=0pt,colframe=red!20, colback=red!20]{Kidney stones $\to$ Laboratory tests $\to$ Serum calcium $\leftarrow $ Laboratory tests $\leftarrow $ Gastric ulcer...
        } (\textbf{example})
           \vspace{5pt}
        
        \textbf{\#\#\# \{\{ User Query \}\}}
        \end{flushleft}
    \end{minipage}
};
\node[fancytitle, rounded corners, right=10pt] at (box.north west) {The Prompt Format of LLM Reader ($\mathcal{P}_{\texttt{Reader}}$)};
\label{tab:prompts}
\end{tikzpicture}%
\end{tabular}
\end{figure*}

\subsubsection{NER Module}
Although there still remains a possibility of an inaccurate comprehension of relationships within $\mathcal{HO}$ (i.e. hallucinations or misunderstanding between medical entities), training a discriminative model or using other general-domain LLMs for authenticity $\mathcal{HO}$ is extremely labor-intensive and will lead to error accumulation. To tackle this issue, we extract entities instead of relationships, and utilize the completely unmistakable triplets in KGs for authenticity instead of the relations analyzed in $\mathcal{HO}$. 
As a consequence, we have trained a medical Named Entity Recognition (NER) model using the CMEEE dataset\footnote{https://tianchi.aliyun.com/dataset/144495}~\cite{cmeee1,cmeee2}. Our NER Module is built upon the W2NER model~\cite{w2ner}, the state-of-the-art word-word NER model that effectively addresses three primary types of NER situations (flat, overlapped, discontinuous). This medical NER model can wonderfully extract medical entities from complex medical contexts:
\begin{equation}
\mathcal{U} = [u_1,\cdots, u_{|\mathcal{U}|}] = \texttt{NER}(\mathcal{Q}\oplus\mathcal{HO}),
    \label{eq:ner}
\end{equation}
where $\oplus$ is the concatenation function and $u_i$ represents the corresponding extracted entity.  

\subsection{Knowledge Graph Retrieval Module}
\label{sec: kg retrieval}
% 讲清楚故事，为什么要用这三种路径？分别有什么用？之前都用的一种，又是为什么？参考CRF
\subsubsection{Embedding Alignment}
Subsequently, we link the potential entity to $\mathcal{KG}$ using dense retrieval methods. This process involves employing an encoding model, denoted as $\texttt{enc}(\cdot)$, to encode the potential entity ${u_i}$ and entities $\mathcal{E}$ within $\mathcal{KG}$. To be specific, we utilize the GTE embedding model~\cite{gte} "gte\_sentence-embedding"\footnote{https://www.modelscope.cn/models/damo/nlp\_gte\_sentence-embedding}, which is currently the top-performing model for text vector embedding in the retrieval field. GTE Encoder follows a two-stage training process: initially using a large-scale dataset with weak supervision from text pairs, followed by fine-tuning with high-quality manually labeled data using Contrastive Learning~\cite{Le_Khac_2020,SPGCL}. 

Then, the inner product similarity between the embeddings of ${u_i}$ and $\mathcal{E}$ is then computed. The entity with the highest similarity, surpassing a predefined threshold $\delta\in[0,1]$, is considered a match. This linkage process can be formulated as follows:
\begin{equation}
\begin{aligned}
& \texttt{sim}({u_i}, e_j) = \bigl \langle \texttt{enc}({u_i}), \texttt{enc}(e_j) \bigl\rangle, \quad u_i \in \mathcal{U}, e_j \in \mathcal{E},
\\
& {u_i} \leftrightarrow e_j \texttt{ iff }  e_j = \{
\mathop{\texttt{argmax} }\limits_{e_k \in \mathcal{E}}\texttt{sim}({u_i}, e_k) ~|~ \texttt{sim}({u_i}, e_j) > \delta\},
\label{eq:kgei}
\end{aligned}
\end{equation}
where $\delta \in [0,1]$ is the threshold hyper-parameter. We utilize the same encoding model $\texttt{enc}(\cdot)$ to embed each medical entity, and $\bigl \langle \texttt{enc}({u_i}), \texttt{enc}(e_j) \bigl\rangle$ denotes the inner product between extracted entities and KGs entities for achieving graph entity linking. Finally, the matched entities set is denoted as $\mathcal{E}_{Q}$.

\subsubsection{Search Reasoning Chains in Knowledge Graph} 
Next, using matched entities, we explore reasoning chains within $k$ hops and consolidate this knowledge along with descriptions of the head and tail entities.
Considering various knowledge graph retrieval methods, we opt for utilizing reasoning chains between entities for several reasons: \textbf{\underline{i)}} Reasoning Chains provide richer logical knowledge provided for LLMs to help it digest, compared to entities and entity descriptions alone. \textbf{\underline{ii)}} Reasoning chains help LLM Reader understand the relationships between different entities, thereby alleviating hallucinations and error problems. \textbf{\underline{iii)}} Reasoning chains act as an efficient pruning mechanism, filtering out noise more effectively than subgraphs and saving token resources. 
% Additionally, paths organize knowledge into reasoning chains, effectively engaging the large model's reasoning capability and deepening its understanding.
% Next, leveraging matched entities, we seek reasoning chains within $k$ hops of each other and integrate the knowledge of these paths in the form of 'entity name - relationship name - entity name.', coupled with the descriptions of the initial and final entities, serves as the knowledge base for retrieval.
% As there are so many knowledge graph retrieval methods, we consider using knowledge paths between entities. The reasons are as follows: compared to simply collecting graph entities and descriptive knowledge, collecting paths can provide more knowledge on $\mathcal{KG}$; compared to using subgraphs, paths can filter out more irrelevant noise, akin to a pruning operation, thus saving token resources more effectively; compared to knowledge from overlapping subgraphs, paths can organize knowledge in the form of reasoning chains, fully stimulating the reasoning capability of the large model and deepening its understanding of knowledge.

As a consequence, in light of~\cite{10.1145/3511808.3557313}, we consider three possible reasoning chains from medical perspective: \textbf{\underline{i)}} \textit{\textbf{Path}} (head-to-tail) as $\text{path}_{ij}$, for comprehensively analyzing the triggering and causal relationships between diseases and symptoms~\cite{path_1,path2}. \textbf{\underline{ii)}} \textit{\textbf{Co-ancestor chain}} (tail-to-tail) as $\text{chainCA}_{ij}$, for referring similar physiological or environmental factors for better analogical diagnosis~\cite{co_anc_1}. \textbf{\underline{iii)}} \textit{\textbf{Co-occurance chain}} (head-to-head) as $\text{chainCO}_{ij}$, for better capturing the pathological characteristics and evolution of diseases~\cite{co-occurance}.
In general, the reasoning chain set $\mathcal{RC}$ after the graph retrieval are as:
\begin{equation}
\footnotesize
\begin{aligned}
& \text{path}_{ij} = (\underbrace{e_i \to r_{\cdot} \to e_{\cdot} \to \cdots \to r_{\cdot} \to e_j}_{\text{within } k \text{ hops}}, d_i, d_j),
\\
& \text{chainCA}_{ij} = (\underbrace{e_i \to r_{\cdot} \to e_{\cdot} \leftarrow \cdots \leftarrow r_{\cdot} \leftarrow e_j}_{\text{within } k \text{ hops}}, d_i,d_j),
\\
& \text{chainCO}_{ij} = (\underbrace{e_i \leftarrow r_{\cdot} \leftarrow e_{\cdot} \to \cdots \to r_{\cdot} \to e_j}_{\text{within } k \text{ hops}}, d_i, d_j), 
\label{eq:reasoning chains}
\end{aligned}
\end{equation}
where $e_i, e_j \in \mathcal{E}_{{Q}}$, and $r_.$ is the relation. For any entity pair in $\mathcal{E_{Q}}$, we collect its reasoning chains within $k (k \geq 2)$ hops and description of head and tail entity $d_i,d_j$ in $\mathcal{KG}$. 

\subsection{Post-Retrieval Phase}
% 讲清楚，为什么要用Chunk-based？因为密度不均衡。会筛除很多有效信息，所以用chunk和HO。
Through retrieval, a large amount of reasoning chains will be collected. However, due to the considerable noise and the shortage of token resources (c.f. Challenge III in Section \ref{intro}), we employ a {reranker model} to prune and eliminate irrelevant noise knowledge by reranking reasoning chains, leading to more efficient token resource utilization. For the reranker base model, we use the "bge\_reranker\_large"\footnote{https://huggingface.co/BAAI/bge-reranker-large}~\cite{bge_embedding}, trained through large-scale text pairs with asymmetric instruction tunning, to map text to a low-dimensional dense vector to rerank $topK$ documents.

Moreover, due to the varying knowledge densities between queries and reasoning chains, traditional re-ranking based solely on $\mathcal{Q}$ may filter out valuable knowledge acquired through \texttt{HOM}, resulting in a repetitive and monotonous situation. As a consequence, we innovatively combine $\mathcal{HO}$ and $\mathcal{Q}$, rather than relying solely on user query, utilizing the richer medical knowledge contained in  $\mathcal{HO}$. Practically, we first remove stop words from natural language and then 
we use the chunk method to segment $\mathcal{HO}$ and $\mathcal{Q}$:
\begin{equation}
 \{\mathcal{C}\}= \texttt{Chunk}(\mathcal{Q}\oplus\mathcal{HO}),
    \label{eq:chunk}
\end{equation}
where $\{\mathcal{C}\}=\{c_1,\cdots, c_i, \cdots, c_{|\{\mathcal{C}\}|}\}$ is the segmented fragments, with carefully selected chunk window size $lc$ and overlap size $oc$. Then, we leverage a reranking model denoted as $\texttt{Rerank}(x, y; topK)$, which means referring to segment set $x$, we select the $topK$ reranked retrieved chains from set $y$. Acting as a filter, the reranking model reevaluates the significance of each chain, considering various factors such as relevance, coherence, and informativeness. 
% In this context, we employ the reranking model to selectively prune and prioritize the $topK$ paths, thereby presenting a more focused and pertinent set of chains as the pruned knowledge. 
% Moreover, we will also chunk $\mathcal{HO}$ and then use the segmented chunks to perform reranking with the reasoning chain, selecting the $topK$ reasoning chain as the filtered results. To avoid the loss of semantic information in the user query, we combine the segmented $\mathcal{HO}$ and $\mathcal{Q}$ list as: $\text{Chunk}([\mathcal{Q}||\mathcal{HO}])$.
The filtering process can be denoted as:
\begin{equation}
\mathcal{RC}_{\texttt{prune}} = \texttt{Rerank}\bigl(\mathcal{RC}, \{\mathcal{C}\}; topK\bigl),
    \label{eq:graph reranking}
\end{equation}
where $|\mathcal{RC}_{\texttt{prune}}|=topK$. 

\subsection{LLM Reader}
% 这个就把prompt说一下就行。
Finally, we link $\mathcal{RC}_{\texttt{prune}}$ with directed arrows, combined with the description of the head and tail entities, and feed the retrieved knowledge as well as user query $\mathcal{Q}$ to LLM Reader via prompt engineering. The prompt format $\mathcal{P}_{\texttt{Reader}}$ is illustrated in Figure \ref{tab:prompts} (Down.) and the LLM's answer can be expressed as:% Practically, we concatenate the user query $\mathcal{Q}$ with the filtered retrieved graph knowledge $\mathcal{RC}_{\text{prune}}$ and input it into the large model to obtain results:
\begin{equation}
\texttt{Answer} = \texttt{LLM}(\mathcal{Q}, \mathcal{RC}_{\texttt{prune}}~|~\mathcal{P}_{\texttt{Reader}}).
    \label{eq:out}
\end{equation}

\subsection{HyKGE Process}
% 把流程捋清楚就行
\label{sec:porcess}
Algorithm \ref{alg:training} shows the overall RAG process of HyKGE.
Given the knowledge graph $\mathcal{KG}$, HyKGE first pre-embed the entity name using $\texttt{enc}(\cdot)$, and saves the vector locally (Lines 1-3). Then, we query LLM to obtain $\mathcal{HO}$ in response to user query $\mathcal{Q}$ (Line 4). After that, we extract entities from $\mathcal{HO}$ and $\mathcal{Q}$ (Line 5) and match them with $\mathcal{KG}$ (Line 6). HyKGE then retrieves the reasoning chains from $\mathcal{KG}$ (Line 7) while filtering the noise path with the HO Fragment Granularity-aware rerank module (Line 8). At last, HyKGE organizes the retrieved knowledge and query via prompt (Line 9) and queries LLM Reader to get the optimized answer (Line 10).
% \vspace{-0.3cm}
\begin{algorithm}[ht]
    \caption{The RAG process of HyKGE.}
    \footnotesize
    \label{alg:training}
    \renewcommand{\algorithmicensure}{\textbf{Output:}}
    \begin{algorithmic}[1]
    \Require Knowledge Graph $\mathcal{KG} = (\mathcal{E}, \mathcal{R}, \mathcal{T}, \mathcal{D}, \mathcal{N})$, token vocabulary set $\mathcal{V}$, user query $\mathcal{Q}$, trained NER model $\texttt{NER}(\cdot)$, trained embedded model $\texttt{enc}(\cdot)$, trained Reranking model $\texttt{Rerank}(\cdot)$, Large Language Model $\texttt{LLM}(\cdot)$, hyper-parameters $\delta, k, topK$. 
    % \State Fill $X$ with the column-wise mean value. \Comment{Data Pre-processing}
    % \State
    \For{$e_i \in \mathcal{E}$}
    \Comment{\textbf{Embed Knowledge Graph}}
    \State Save $\texttt{enc}(e_i)$ locally;
    \EndFor
    \State Obtain $\mathcal{HO}$ via LLM: $\mathcal{HO}=\texttt{LLM}(\mathcal{Q}~|~\mathcal{P}_{\texttt{HO}})$;    \Comment{\textbf{Hypothesis Output}}
    \State Extract entities $\mathcal{U}$ from $\mathcal{HO}$ and $\mathcal{Q}$ via Eq.\eqref{eq:ner};   \Comment{\textbf{NER Module}}
    \State Match $\mathcal{U}$ with $\mathcal{E}$ via Eq.\eqref{eq:kgei} and attain $\mathcal{E}_Q$;   \Comment{\textbf{Entity Linking}} 
    \State Retrieved reasoning chains between any two anchor entities from $\mathcal{KG}$; \Comment{\textbf{Knowledge Graph Retrieval}}
    \State Filter noise reasoning chains with HO Fragment Granularity-aware Rerank Module; \Comment{\textbf{Prune knowledge}}
    \State Organize knowledge $\mathcal{RC}_{\texttt{prune}}$ with user query $\mathcal{Q}$ into prompt;
    \State Get optimized answer of LLMs; \Comment{\textbf{LLM Reader}}
    \end{algorithmic}
\end{algorithm}
\vspace{-0.2cm}

\section{Experiments}
In this section, we conduct a series of experiments on two datasets to answer the following research questions:
\begin{itemize}[leftmargin=*]
    \item \textbf{RQ1} (Section \ref{rq1}): Does HyKGE outperform the state-of-the-art Knowledge Graph RAG methods using the same database source?
    \item \textbf{RQ2} (Section \ref{Ablation Study (RQ 2)}, \ref{rq23}, \ref{subsection:Component Efficiency Analysis}): Is the framework we designed effective? What impact does each module have on the overall performance?
    \item \textbf{RQ3} (Section \ref{rq3}, \ref{rq23}): Does the retrieved knowledge we provide enhance the interpretability of LLMs answers?
    \item \textbf{RQ4} (Section \ref{rq4}): How sensitive is HyKGE to hyper-parameters retrieval hop $k$ and rerank threshold $topK$?
\end{itemize}

\subsection{Experimental Setup}
\begin{table}[t]
\vspace{-0.2cm}
  \centering
    \footnotesize
   \setlength\tabcolsep{2pt}
  \caption{Analysis Comparison of RAG methods. Average Duration is computed based on GPT 3.5 turbo.}
  \vspace{-0.2cm}
  \label{table1}
  \resizebox{1\linewidth}{!}{
  \begin{tabular}{c|c|c|c|c}
    \hline
    \rowcolor[gray]{0.9}
     & \textbf{External} & \multicolumn{2}{c|}{\textbf{LLMs RAG Opt.}}  & \textbf{Avg.}\\
    \cline{3-4}
    \rowcolor[gray]{0.9}
    \textbf{Method}& \textbf{Knowledge}  & \textbf{Finetuning Retriever} & \textbf{LLMs Interactions (/times)}  & \textbf{Time~(s)}\\
    Base & \xmark  & \xmark  & $1$ & $7.42$ \\
    KGRAG & \cmark & \xmark & $1$ & $13.28$ \\
    QE & \cmark & \xmark  & $2$ & $18.54$ \\
    CoN & \cmark  & \cmark & $\geq 2$ & $34.33$ \\
    CoK & \cmark & \xmark & $\geq 4$ & $45.84$ \\
    KALMV & \cmark & \xmark & $\geq 4$ & $47.23$ \\
    KG-GPT & \cmark & \xmark & $5$ & $55.19$ \\
    SuRE & \cmark & \xmark & $\geq 5$ & $63.08$ \\
    \hdashline 
    \textbf{HyKGE (ours)} & \cmark  & \xmark & $2$  & $19.76$\\
    \hline
  \end{tabular}}
  \vspace{-0.2cm}
\end{table}
\begin{table*}[!t]
\caption{Performance comparison (in percent ± standard deviation) on CMB-Exam and MMCU-Medical for medical Q\&A answer. \redfont{Red shading} indicates the best-performing model, while \bluefont{blue} signifies the second-best in the ablation study, and \greenfont{green} signifies the second-best in baselines. }
\setlength\tabcolsep{1.6pt}
\vspace{-0.3cm}
\footnotesize
\label{tab:comparison}
\centering
\resizebox{1.0\linewidth}{!}{
\begin{tabular}{cc|cc|cc|cc|cc}
\hline
\textbf{LLM Turbo} & \multicolumn{1}{c|}{\textbf{LLM}} & \multicolumn{4}{c|}{GPT 3.5} & \multicolumn{4}{c}{Baichuan 13B-Chat} \\
\hline
\multirow{2}{*}{\textbf{Method}} & \multicolumn{1}{c|}{\textbf{Dataset}} & \multicolumn{2}{c|}{MMCU-Medical} & \multicolumn{2}{c|}{CMB-Exam} & \multicolumn{2}{c|}{MMCU-Medical} & \multicolumn{2}{c}{CMB-Exam} \\
\cline{2-10}
& \textbf{Metric} & EM & PCR & EM & PCR & EM & PCR & EM & PCR \\
\hline
\multirow{8}{*}{\textbf{Baselines}} & Base & 43.52$\pm$1.92 & 50.55$\pm$1.88 & 38.40$\pm$2.03 & 46.76$\pm$1.93 & 42.20$\pm$2.87 & 46.09$\pm$2.65 & 36.91$\pm$2.94 & 40.95$\pm$2.70 \\
\cdashline{2-10}
& KGRAG & 38.74$\pm$1.66 & 43.38$\pm$1.68 & 38.00$\pm$1.90 & 42.26$\pm$1.88 & 34.37$\pm$2.36 & 38.51$\pm$2.10 & 39.92$\pm$2.37 & 45.84$\pm$2.29 \\
& QE & 40.28$\pm$1.15 & 46.79$\pm$1.41 & 36.35$\pm$0.88 & 41.84$\pm$1.10 & 38.25$\pm$2.23&  44.23$\pm$1.94 & 34.27$\pm$2.88 &38.79$\pm$2.65 \\
& CoN & \cellcolor{green!15}{45.74$\pm$1.42} & 51.15$\pm$1.94 & \cellcolor{green!15}{42.45$\pm$1.06}& 45.65$\pm$1.65 &44.98$\pm$2.65 &50.65$\pm$1.94 &41.37$\pm$2.45 &47.58$\pm$2.73 \\
& CoK & 45.15$\pm$1.59 & \cellcolor{green!15}{52.35$\pm$1.77} & 42.32$\pm$1.35&\cellcolor{green!15}{45.98$\pm$1.80} &\cellcolor{green!15}{45.15$\pm$1.86} & \cellcolor{green!15}{51.19$\pm$1.69}&\cellcolor{green!15}{41.87$\pm$2.18} &\cellcolor{green!15}{47.95$\pm$1.79} \\
% \cdashline{1-10}
& KALMV & 39.24$\pm$1.41 & 43.77$\pm$1.23 & 38.24$\pm$0.84&43.37$\pm$1.89 &36.17$\pm$2.33 & 40.85$\pm$2.11&38.61$\pm$2.44 &43.92$\pm$1.97 \\
& KG-GPT & 45.08$\pm$1.96 & 52.16$\pm$1.54 & 41.49$\pm$1.04 &45.72$\pm$1.48 &44.25$\pm$2.38 & 50.97$\pm$2.65& 39.92$\pm$2.38 & 45.20$\pm$1.49 \\
& SuRe & 44.81$\pm$1.38 & 51.49$\pm$1.97 & 41.37$\pm$1.26 & 44.27$\pm$1.47 &44.77$\pm$1.80 & 50.24$\pm$2.09 & 39.49$\pm$1.57 &46.22$\pm$1.70\\
\cdashline{1-10}
\textbf{Ours} & \textbf{HyKGE} & \cellcolor{red!20}{\textbf{49.65$\pm$1.39}} & \cellcolor{red!20}{\textbf{57.82$\pm$1.54}} & \cellcolor{red!20}{\textbf{45.94$\pm$1.20}} & \cellcolor{red!20}{\textbf{50.63$\pm$1.33}} & \cellcolor{red!20}{\textbf{49.33$\pm$1.72}} & \cellcolor{red!20}{\textbf{58.12$\pm$1.79}} & \cellcolor{red!20}{\textbf{45.44$\pm$1.97}} & \cellcolor{red!20}{\textbf{51.25$\pm$1.84}} \\
\hline
\rowcolor[gray]{0.95}
\multicolumn{2}{c|}{*\textbf{Performance Gain $\uparrow$}} & 8.55$\sim$28.16 & 10.45$\sim$33.29 & 8.38$\sim$26.38& 8.28$\sim$21.01  &9.26$\sim$43.53& 13.54$\sim$50.92& 8.53$\sim$32.59& 6.88$\sim$32.12\\
\hline
\multirow{5}{*}{\textbf{Ablation}} & HyKGE \texttt{(w/o HO)} & 41.08$\pm$1.45 & 49.74$\pm$1.84 & 34.40$\pm$1.13 &40.14$\pm$1.25 & 39.55$\pm$1.98 & 45.28$\pm$2.14 & 33.33$\pm$2.22 & 35.42$\pm$2.70\\
& HyKGE \texttt{(w/o Chains)} & 48.15$\pm$1.75 & \cellcolor{cyan!20}{54.53$\pm$1.68} & 44.60$\pm$0.94 & 48.27$\pm$1.04 & \cellcolor{cyan!20}{48.65$\pm$1.91} & \cellcolor{cyan!20}{55.45$\pm$1.81} & 43.40$\pm$1.80& 48.81$\pm$2.75\\
& HyKGE \texttt{(w/o Description)} & \cellcolor{cyan!20}{48.30$\pm$1.45} & 54.01$\pm$1.86 & \cellcolor{cyan!20}{44.80$\pm$1.33} & \cellcolor{cyan!20}{48.56$\pm$1.41} & 48.22$\pm$2.12 & 55.23$\pm$1.86 &43.77$\pm$2.37 &\cellcolor{cyan!20}{49.86$\pm$1.47} \\
& HyKGE \texttt{(w/o Fragment)} & 47.87$\pm$1.66 & 54.34$\pm$1.49 & 42.33$\pm$1.02 & 47.54$\pm$0.84&47.95$\pm$1.90& 53.45$\pm$2.33&\cellcolor{cyan!20}{44.72$\pm$2.66} &49.29$\pm$2.56 \\
& HyKGE \texttt{(w/o Reranker)} & 46.38$\pm$1.65 & 52.48$\pm$1.88 & 41.44$\pm$0.88 & 48.84$\pm$1.09& 43.59$\pm$2.34&46.88$\pm$2.56&40.65$\pm$2.27 &46.25$\pm$2.11 \\
\hline
\end{tabular}}
\end{table*}

\begin{table*}[!t]
\caption{RAG relevance and answer performance comparison (in mean ± standard deviation) on CMB-Exam, MMCU-Medical and CMB-Clin for medical Q\&A answer with GPT 3.5 Turbo. }
\setlength\tabcolsep{1.5pt}   
% \redfont{Red shading} indicates the best-performing model, while \bluefont{blue} signifies the second-best in the ablation study, and \greenfont{green} signifies the second-best in baselines. }
\vspace{-0.3cm}
\footnotesize
\label{tab:comparison2}
\centering
\resizebox{1.0\linewidth}{!}{
\begin{tabular}{cc|ccc|ccc|cccc}
\hline
\multirow{2}{*}{\textbf{Method}} & \multicolumn{1}{c|}{\textbf{Dataset}} & \multicolumn{3}{c|}{MMCU-Medical} & \multicolumn{3}{c|}{CMB-Exam} & \multicolumn{4}{c}{CMB-Clin} \\
\cline{2-12}
& \textbf{Metric} & ACJ & PPL & ROUGE-R  & ACJ & PPL & ROUGE-R & BLEU-1 & BLEU-4 & PPL & ROUGE-R\\
\hline
\multirow{8}{*}{\textbf{Baselines}} 
& Base & / & 47.42$\pm$1.24 & / & / & 62.54$\pm$0.94  & / &4.83$\pm$1.21 &6.51$\pm$1.55 & 10.38$\pm$1.47 & 23.99$\pm$1.06 \\ \cdashline{2-12}
& KGRAG &13.38$\pm$4.27 & 151.22$\pm$2.87 & {5.31$\pm$0.97} & 18.40$\pm$5.58 &{218.67$\pm$3.68} &{11.25$\pm$1.93} & 5.34$\pm$1.51 & 8.77$\pm$1.90 &61.81$\pm$2.51 & 22.15$\pm$1.27  \\
& QE &{25.53$\pm$3.68} & {28.75$\pm$1.58} &14.05$\pm$1.22 &{31.91$\pm$6.82} &29.57$\pm$1.60 &16.64$\pm$2.11 & 8.85$\pm$1.97& 18.67$\pm$1.44 &28.32$\pm$2.48 & 26.24$\pm$2.20  \\
& CoN &19.14$\pm$5.18 &29.01$\pm$1.61 & 16.46$\pm$1.19& 14.89$\pm$5.53&{27.35$\pm$1.93} &17.31$\pm$1.48 & 12.48$\pm$1.65 & 25.81$\pm$1.04 &17.65$\pm$3.47 & \cellcolor{green!15}{31.37$\pm$1.87} \\
& CoK &18.45$\pm$4.71 &\cellcolor{green!15}{24.38$\pm$1.93} & \cellcolor{green!15}{18.23$\pm$2.02}&16.77$\pm$6.71 &28.69$\pm$2.26 &\cellcolor{green!15}{19.94$\pm$1.46}& 12.35$\pm$1.46 & 24.79$\pm$1.18 &21.57$\pm$2.62  & 30.86$\pm$2.24 \\
& KALMV & 14.42$\pm$3.88 & 147.22$\pm$3.12 & 7.21$\pm$1.08&18.77$\pm$5.91 &233.49$\pm$4.19 & 12.84$\pm$1.34&5.72$\pm$1.16 &8.27$\pm$1.20 & 80.46$\pm$2.51 & 23.16$\pm$2.23\\
& KG-GPT & \cellcolor{green!15}{32.03$\pm$4.82} & 25.76$\pm$2.45 & 15.90$\pm$1.31 & \cellcolor{green!15}{38.70$\pm$5.44} & \cellcolor{green!15}{24.01$\pm$3.96} & 17.72$\pm$1.80 & \cellcolor{green!15}{13.03$\pm$0.76} & \cellcolor{green!15}{26.14$\pm$1.09} & \cellcolor{green!15}{15.54$\pm$1.38} & 28.42$\pm$1.91\\
& SuRe & 20.16$\pm$3.93 & 26.49$\pm$2.88 & 16.91$\pm$1.84 &22.27$\pm$4.02 &30.81$\pm$2.59 & 16.18$\pm$1.70 &10.54$\pm$0.92 &24.82$\pm$1.31 & 16.84$\pm$1.46 & 29.18$\pm$1.62\\
\cdashline{1-12}
\textbf{Ours} & \textbf{HyKGE} & \cellcolor{red!20}\textbf{59.57$\pm$4.37}& \cellcolor{red!20}\textbf{12.55$\pm$1.29}& \cellcolor{red!20}\textbf{26.89$\pm$1.67} & \cellcolor{red!20}\textbf{71.28$\pm$3.88} & \cellcolor{red!20}\textbf{10.14$\pm$1.68} & \cellcolor{red!20}\textbf{32.11$\pm$1.28} &\cellcolor{red!20}\textbf{18.28$\pm$0.48}&\cellcolor{red!20}\textbf{30.21$\pm$1.05}&
\cellcolor{red!20}\textbf{8.56$\pm$1.24}  & \cellcolor{red!20}\textbf{33.66$\pm$1.54} \\
\hline
\rowcolor[gray]{0.95}
\multicolumn{2}{c|}{*\textbf{Performance Gain}} & 133.33$\sim$345.22 & 48.52$\sim$91.70 & 45.75$\sim$406.40 &84.19$\sim$378.71 & 57.77$\sim$95.36 & 61.03$\sim$185.42 &40.29$\sim$278.47&15.57$\sim$364.06&46.85$\sim$89.25 & 7.30$\sim$51.96\\
\hline
\multirow{2}{*}{\textbf{Ablation}} & HyKGE \texttt{(w/o HO)} & \cellcolor{cyan!20}{41.49$\pm$5.36} & \cellcolor{cyan!20}{15.57$\pm$2.31} & 22.30$\pm$2.37& \cellcolor{cyan!20}{51.48$\pm$4.92} &\cellcolor{cyan!20}{11.23$\pm$1.96} &\cellcolor{cyan!20}{29.01$\pm$1.96} &7.15$\pm$2.35&11.55$\pm$1.89&\cellcolor{cyan!20}{8.96$\pm$1.01} & 30.48$\pm$2.58 \\
& HyKGE \texttt{(w/o Fragment)} & 38.30$\pm$4.85 & 18.95$\pm$2.04 & \cellcolor{cyan!20}{23.63$\pm$1.47}&41.91$\pm$4.44 &{11.26$\pm$1.45} &26.89$\pm$2.65 &\cellcolor{cyan!20}{11.28$\pm$1.76}&\cellcolor{cyan!20}{23.09$\pm$1.44}&8.99$\pm$1.72 & \cellcolor{cyan!20}{31.40$\pm$0.82} \\
\hline
\end{tabular}}
\vspace{-0.3cm}
\end{table*}

\begin{table}[!t]
\caption{Performance and computation time comparison (in mean ± standard deviation) on MMCU-Medical for medical Q\&A answer with GPT 3.5 Turbo. }
\setlength\tabcolsep{1.4pt}
% \redfont{Red shading} indicates the best-performing model, and \greenfont{green} signifies the second-best in baselines. }
\vspace{-0.3cm}
\footnotesize
\label{tab:comparison3}
\centering
\resizebox{1\linewidth}{!}{
\begin{tabular}{c|ccc}
\hline
\textbf{Method~/~Metric} & EM & PCR  &Avg. Time~(s)\\
\hline
HyKGE & \cellcolor{red!20}\textbf{49.65$\pm$1.39} & \cellcolor{red!20}\textbf{57.82$\pm$1.54} & \cellcolor{red!20}\textbf{19.76}\\
HyKGE($+$ LLM for NER) & \cellcolor{green!15}{48.17$\pm$1.13} & \cellcolor{green!15}{56.77$\pm$1.02} & \cellcolor{green!15}{26.61}\\
HyKGE($+$ LLM for Reranker) & {42.72$\pm$2.06} & {48.24$\pm$1.17} & 32.51\\
HyKGE($+$ LLM for Summary) & {43.02$\pm$3.11} & {46.54$\pm$2.08} & 28.51\\
\hline
\end{tabular}}
\vspace{-0.3cm}
\end{table}

\subsubsection{Dataset.}
Our experiments are conducted on two open-source query sets: MMCU-Medical~\cite{mmcu} and CMB-Exam~\cite{cmb} datasets, which are designed for multi-task Q\&A and encompass single and multiple-choice questions in the medical field,
and one open-domain Q\&A dataset CMB-Clin~\cite{cmb} which is the inaugural multi-round question-answering dataset based on real, complex medical diagnosis and treatment records. 
For MMCU-Medical, the questions are from the university medical professional examination, covering the three basic medical sciences, pharmacology, nursing, pathology, clinical medicine, infectious diseases, surgery, anatomy, etc., with a total of 2,819 questions.
The CMB-Exam dataset utilizes qualifying exams as a data source in the four clinical medicine specialties of physicians, nurses, medical technicians, and pharmacists, with a total of 269,359 questions. Given the extensive size of the CMB-Exam dataset, we randomly sample 4,000 questions for testing. 
The CMB-Clin dataset contains 74 high-quality, complex, and real patient cases with 208 medical questions.

\subsubsection{Knowledge Graph.}
\textit{CMeKG} (Clinical Medicine Knowledge Graph)\footnote{https://cmekg.pcl.ac.cn/, https://github.com/king-yyf/CMeKG\_tools} ~\cite{cmekg}, \textit{CPubMed-KG} (Large-scale Chinese Open Medical Knowledge Graph) \footnote{https://cpubmed.openi.org.cn/graph/wiki} and \textit{Disease-KG} (Chinese disease Knowledge Graph)\footnote{https://github.com/nuolade/disease-kb} are open-source medical KGs, which integrates extensive medical text data, including diseases, medications, symptoms and diagnostic treatment technologies. 
The fused KG has 1,288,721 entities and 3,569,427 relations. However, due to the lack of medical entity descriptions in its entities, we collect relevant entity knowledge from Wikipedia\footnote{https://www.wikipedia.org/}, Baidu Baike\footnote{https://baike.baidu.com/}, and Medical Baike\footnote{https://www.yixue.com/}, and store them as entity descriptions.

\subsubsection{LLM Turbo.}
To fairly verify whether HyKGE can effectively enhance LLMs, we selected the following two types of general-domain large models as the base model and explored the gains brought by HyKGE: GPT 3.5 and Baichuan13B-chat~\cite{baichuan}.
\subsubsection{Compared Methods.}
In order to explore the advantages of the HyKGE, we compare the HyKGE results against eight other models: 
(1) \textbf{Base Model (Base)} servers as the model without any external knowledge, used to check the improvement effect of different RAG methods. We use GPT 3.5 and Baichuan13B-chat as base models.
(2) \textbf{Knowledge Graph Retrieval-Augmented Generation (KGRAG)~\cite{soman2023biomedical,kgrag_arxiv,Sen2023KnowledgeGL}} uses user query as a reference to retrieve in the KGs, which is the base model of RAG on KG and has been widely applied in~\cite{soman2023biomedical,kgrag_arxiv,Sen2023KnowledgeGL}.
(3) \textbf{Query Expansion (QE)~\cite{qe}} reformulate the user's initial query by adding additional terms with a similar meaning with the help of LLMs.
(4) \textbf{CHAIN-OF-NOTE (CoN)~\cite{con}} generates sequential reading notes for retrieved knowledge, enabling a thorough evaluation of their relevance to the given question and integrating these notes to formulate the final answer.
(5) \textbf{Chain-of-Knowledge (CoK)~\cite{cok}} utilize the power of LLMs and consists of reasoning preparation, dynamic knowledge adapting, and answer consolidation. (6) \textbf{Knowledge-Augmented Language Model Verification (KALMV)~\cite{baek2023knowledgeaugmented}} verifies the output and the knowledge of the knowledge-augmented LLMs with a separate verifier.
(7) \textbf{Knowledge Graph Generative Pre-Training (KG-GPT)~\cite{kim2023kggpt}} comprises three steps: Sentence Segmentation, Graph Retrieval, and Inference, each aimed at partitioning sentences, retrieving relevant graph components, and deriving logical conclusions. (8) \textbf{Summarizing Retrievals  (SuRe)~\cite{chen2024improving}} constructs summaries of the retrieved passages for each of the multiple answer candidates and confirms the most plausible answer from the candidate set by evaluating the validity and ranking of the generated summaries. Note that we follow the prompts of the baselines as stated strictly.
The baselines and running time are summarized in Table~\ref{table1}. 
In RAG Options, CoN requires fine-tuning the retriever, implying a higher training overhead and the prerequisite of preparing a dataset. In addition, it is also difficult to migrate to other domain-specific KGs. In terms of LLMs interactions, QE, CoN, CoK, KALMV, KG-GPT, SuRe and HyKGE all necessitate engagement with LLMs. However, CoN, CoK, KALMV, KALMV, KG-GPT and SuRe entail multiple interactions (more than twice), significantly escalating the time expenditure.

\subsubsection{Evaluation Metrics.}
As for the evaluation of multi-task medical choice question performance, we guide LLMs to only answer the correct answer and employ established metric \textbf{Exact Match (EM)} as suggested by prior work~\cite{f1,f2}. For the EM score, an answer is deemed acceptable if its form corresponds to all correct answers in the provided list. For multiple-choice questions, we also calculate a \textbf{Partial Correct Rate (PCR)}. 
In comparison to EM, if there is a missing answer without any incorrect ones, PCR classifies it as correct.
In addition, to verify the effectiveness of the retrieved knowledge, we also let LLMs output a complete analysis process. Then, we measure \textbf{Artificial Correlation Judgement (ACJ)} by inviting 20 medical experts to rate the retrieved knowledge according to the criteria of (correlation=1, relevant but useless=0, irrelevant=-1), and calculate the relevant scores for each question by sampling 100 questions from the two datasets. Moreover, we also objectively evaluated the \textbf{Perplexity (PPL)} of LLMs output. The smaller the PPL, the greater the role of retrieved knowledge in reducing LLMs' hallucinations. Moreover, we also complement our analysis with \textbf{ROUGE-Recall (ROUGE-R)}~\cite{xu2023contextaware}. \textbf{ROUGE-R} measures the extent to which the LLMs' responses cover the retrieved knowledge, which is crucial for ensuring comprehensive information coverage. 
For open-domain medical Q\&A tasks, we utilize \textbf{ROUGE-R} and \textbf{Bilingual Evaluation Understudy }(BLEU-1 for answer precision, BLEU-4 for answer fluency) ~\cite{xu2023contextaware} to gauge the similarity of LLMs responses to the ground-truth doctor analysis. Additionally, we employ \textbf{PPL} to assess the quality of LLMs responses.

\subsubsection{Experimental Implementation.}
In HyKGE, $k=3, topK=10, \delta=0.7, lc=10, oc=4$. The prompts for LLMs can refer to Table~\ref{tab:prompts}. {Moreover, for all the baselines and HyKGE, we set the maximum number of returned tokens for LLMs to 500 and the temperature to 0.6.} In all baselines and HyKGE, we first use the Jieba library in Python to perform word segmentation, and then use filtered text to filter out tone words and invalid characters following ``chinese\_word\_cut.txt''\footnote{https://github.com/Robust-Jay/NLP\_Chinese\_WordCut/blob/master/stopwords.txt} to avoid errors in knowledge extraction. 
For a fair comparison, we apply the same W2NER, GTE and FlagEmbedding models for all baselines. Moreover, the parameters of W2NER are optimized with Adam optimizer~\cite{Adam} with $L_2$ regularization and dropout on high-quality medical dataset~\cite{cmeee1,cmeee2}, the learning rate is set to 1e-3, the hidden unit is set to 1024 and weight decay is 1e-4. 
Similar to previous work~\cite{kgrag_arxiv}, because of the randomness of LLMs' outputs, we repeat experiments with different random seeds five times and report the average and standard deviation results. Experimental results are statistically significant
with $p < 0.05$. Implementations are done using the PyTorch 1.9.0 framework~\cite{paszke2019pytorch} in Python 3.9, on an Ubuntu server equipped with 8 A100 GPU and an Intel(R) Xeon(R) CPU.

\subsection{Performance Comparison (RQ 1)}
\label{rq1}
To answer RQ1, we conduct experiments and report results of the accuracy on the MMCU-Medical, CMB-Exam and CMB-Clin datasets with two LLM turbos GPT 3.5 and Baichuan 13B-Chat, as illustrated in Table~\ref{tab:comparison} and Table~\ref{tab:comparison2}. From the reported accuracy, we can find the following observations:

\textbf{Comparison of RAG methods and Base LLMs.} 
Through comparison, we observe that most RAG approaches do not consistently yield effective outcomes when integrated with KGs, especially in contrast with the Base model. For instance, the KGRAG method extracts triples from KG without engaging in essential post-processing steps like reranking and filtering, thereby infusing an overabundance of noise and compromising the interpretative performance of LLMs. As for QE tasks, while traditional QE methods typically show efficacy, LLMs demonstrate a notable difficulty in comprehending instructions that necessitate the task-specific rewriting of multiple-choice questions, which, in turn, detrimentally impacts LLMs performance in such scenarios. Moreover, this effect is particularly pronounced in weaker models, such as Baichuan, where the repercussions of these deficiencies are significantly magnified. However, the improvement in CoN, CoK, KG-GPT, SuRe and HyKGE is more remarkable, because leveraging LLMs to explore or organize knowledge can assist in finding more relational knowledge and the reranking or filtering methods can highly likely remove irrelevant noise knowledge chains, and contribute to accuracy improvement.

\textbf{Comparison of HyKGE and other RAG methods.} Firstly, it is evident that our model, HyKGE, outperforms the baseline models across all metrics.  For instance, the EM and PCR scores see an improvement of approximately \textbf{8.55\%-28.15\%} and \textbf{10.45\%-33.29\%} for the MMCU-Medical dataset with GPT 3.5 turbo, and the BLEU-1 and ROUGE-R scores see an improvement of approximately \textbf{40.29\%-278.47\%} and \textbf{7.30\%-51.69\%} for the CMB-Clin dataset with GPT 3.5 turbo. This highlights the effectiveness of our modules in locating valid information and filtering noises in retrieved knowledge. Although CoK, CoN, KG-GPT and SuRe have achieved commendable results, their advancements are constrained in the knowledge search space, due to their focus on continuous knowledge understanding rather than exploration.
Moreover, compared to CoK, CoN, KG-GPT and SuRe, HyKGE avoids accumulating errors in the chain of thought while acquiring and retaining more relevant yet diverse knowledge. 
{In summary, our proposed HyKGE model exhibits superior performance over all baselines with fewer interaction times with LLMs (c.f. Table~\ref{table1}). Evidenced by comprehension experiments, HyKGE demonstrates the HO Module's and the HO Fragment Granularity-aware rerank module's effectiveness compared to CoN, CoK, KG-GPT and SuRe.}

\begin{figure*}[ht!]
\caption{Case study. We demonstrate the User Query $\mathcal{Q}$, Hypothesis Output $\mathcal{HO}$, Retrieved Reasoning Chains $\mathcal{RC}$ and Pruned Reasoning chains $\mathcal{RC}_{\texttt{prune}}$ of HyKGE using GPT-3.5 Turbo to verify the interpretability and effectiveness of HyKGE. \tcbox[on line, boxsep=1pt, left=0pt,right=0pt,top=0pt,bottom=0pt,colframe=red!30, colback=red!30]{Red shed} signifies that the knowledge or answer is derived from evidence in $\mathcal{Q}$, \tcbox[on line, boxsep=1pt, left=0pt,right=0pt,top=0pt,bottom=0pt,colframe=cyan!20, colback=cyan!20]{blue shed} indicates that the evidence originates from $\mathcal{HO}$, and \tcbox[on line, boxsep=1pt,left=0pt,right=0pt,top=0pt,bottom=0pt,colframe=green!20, colback=green!20]{green shed} represents a corrected answer with the help of $\mathcal{KG}$, despite initially being false in $\mathcal{HO}$.}
\centering
\begin{tabular}{c} 
% Palmyra-Med
\begin{tikzpicture}
\node [mybox] (box){%
    \begin{minipage}{0.935\textwidth}
    \footnotesize
    \setstretch{1.2}
        \begin{flushleft} % Center the content
        {\textbf{\#\#\# User Query $\mathcal{Q}$:}} \\I feel \tcbox[on line, boxsep=1pt, left=0pt,right=0pt,top=0pt,bottom=0pt,colframe=red!30, colback=red!30]{stomach reflux} after eating. What medicine should I take?

        \vspace{5pt}

        {\textbf{\#\#\# Hypothesis Output $\mathcal{HO}$:}} \\ 
        \tcbox[on line, boxsep=1pt, left=0pt,right=0pt,top=0pt,bottom=0pt,colframe=red!30, colback=red!30]{Stomach reflux} can be a symptom caused by \tcbox[on line, boxsep=1pt, left=0pt,right=0pt,top=0pt,bottom=0pt,colframe=cyan!20, colback=cyan!20]{gastroesophageal reflux}. 1. Do you have symptoms related to \tcbox[on line, boxsep=1pt, left=0pt,right=0pt,top=0pt,bottom=0pt,colframe=cyan!20, colback=cyan!20]{indigestion}, \tcbox[on line, boxsep=1pt, left=0pt,right=0pt,top=0pt,bottom=0pt,colframe=cyan!20, colback=cyan!20]{excess stomach acid}, \tcbox[on line, boxsep=1pt, left=0pt,right=0pt,top=0pt,bottom=0pt,colframe=cyan!20, colback=cyan!20]{gastric ulcers}, \tcbox[on line, boxsep=1pt, left=0pt,right=0pt,top=0pt,bottom=0pt,colframe=cyan!20, colback=cyan!20]{bile reflux}, or \tcbox[on line, boxsep=1pt, left=0pt,right=0pt,top=0pt,bottom=0pt,colframe=cyan!20, colback=cyan!20]{stomach pain}? $\ldots$ When experiencing mild gastroesophageal reflux, the following treatment plans can be considered:- Avoid overeating - Avoid spicy, stimulating foods, greasy foods, and alcohol - Maintain a light diet, eat more vegetables. Under appropriate evidence, consider taking \tcbox[on line, boxsep=1pt, left=0pt,right=0pt,top=0pt,bottom=0pt,colframe=cyan!20, colback=cyan!20]{H2 receptor antagonists} or \tcbox[on line, boxsep=1pt, left=0pt,right=0pt,top=0pt,bottom=0pt,colframe=cyan!20, colback=cyan!20]{proton pump inhibitors}. \tcbox[on line, boxsep=1pt,left=0pt,right=0pt,top=0pt,bottom=0pt,colframe=green!20, colback=green!20]{It is not recommended to take} \tcbox[on line, boxsep=1pt, left=0pt,right=0pt,top=0pt,bottom=0pt,colframe=cyan!20, colback=cyan!20]{calcium carbonate} or \tcbox[on line, boxsep=1pt, left=0pt,right=0pt,top=0pt,bottom=0pt,colframe=cyan!20, colback=cyan!20]{aluminum hydroxide} \tcbox[on line, boxsep=1pt,left=0pt,right=0pt,top=0pt,bottom=0pt,colframe=green!20, colback=green!20]{medications, as calcium carbonate can lead to kidney} \tcbox[on line, boxsep=1pt,left=0pt,right=0pt,top=0pt,bottom=0pt,colframe=green!20, colback=green!20]{stones, and the aluminum element in aluminum hydroxide is toxic and may cause allergic reactions or discomfort.}
        
        \vspace{5pt}
        
        {\textbf{\#\#\# Retrieved Reasoning Chains $\mathcal{RC}$:}} \\Kidney stones $\to$ Laboratory tests $\to$ Serum calcium $\leftarrow $ Laboratory tests $\leftarrow $ Gastric ulcer. Kidney stones $\to$ Clinical symptoms and signs $\to$ Fever $\leftarrow $ Complications $\leftarrow $ Esophageal cancer. \tcbox[on line, boxsep=1pt, left=0pt,right=0pt,top=0pt,bottom=0pt,colframe=cyan!20, colback=cyan!20]{Excess gastric acid $\to$ Affected area $\to$ Abdomen $\leftarrow $ Affected area $\leftarrow $ Stomach pain.} Niflumic acid $\to$ Adverse reaction $\to$ Discomfort $\leftarrow $ Adverse reaction $\leftarrow $ Calcium carbonate. Gastrointestinal bleeding $\to$ Related diseases $\to$ Gastric ulcer. \tcbox[on line, boxsep=1pt, left=0pt,right=0pt,top=0pt,bottom=0pt,colframe=cyan!20, colback=cyan!20]{Gastric ulcer $\to$ Treatment $\to$ Aluminum hydroxide.} \tcbox[on line, boxsep=1pt,left=0pt,right=0pt,top=0pt,bottom=0pt,colframe=green!20, colback=green!20]{Calcium carbonate $\to$ Indication $\to$ Excess gastric acid.} \tcbox[on line, boxsep=1pt,left=0pt,right=0pt,top=0pt,bottom=0pt,colframe=green!20, colback=green!20]{Aluminum hydroxide $\to$ Indication $\to$ Gastric reflux.} \tcbox[on line, boxsep=1pt, left=0pt,right=0pt,top=0pt,bottom=0pt,colframe=cyan!20, colback=cyan!20]{Niflumic acid $\to$ Third level classification $\to$ Others $\leftarrow $ Affected area $\leftarrow $ Postprandial food reflux.} \tcbox[on line, boxsep=1pt, left=0pt,right=0pt,top=0pt,bottom=0pt,colframe=cyan!20, colback=cyan!20]{Duodenogastric reflux and bile reflux gastritis $\to$ Treatment $\to$ Calcium carbonate.} More else $\ldots$
       {[125 reasoning chains in total.]

           \vspace{5pt}
        
        {\textbf{\#\#\# Pruned Reasoning chains $\mathcal{RC}_{\texttt{prune}}$:}}

        \tcbox[on line, boxsep=1pt, left=0pt,right=0pt,top=0pt,bottom=0pt,colframe=cyan!20, colback=cyan!20]{Gastric ulcer $\to$ Treatment $\to$ Aluminum hydroxide.} \tcbox[on line, boxsep=1pt,left=0pt,right=0pt,top=0pt,bottom=0pt,colframe=green!20, colback=green!20]{Calcium carbonate $\to$ Indication $\to$ Excess gastric acid.} \tcbox[on line, boxsep=1pt,left=0pt,right=0pt,top=0pt,bottom=0pt,colframe=green!20, colback=green!20]{Aluminum hydroxide $\to$ Indication $\to$ Gastric reflux.} \tcbox[on line, boxsep=1pt, left=0pt,right=0pt,top=0pt,bottom=0pt,colframe=cyan!20, colback=cyan!20]{Niflumic acid $\to$ Third level classification $\to$ Others $\leftarrow $ Affected area $\leftarrow $ Postprandial food reflux.} \tcbox[on line, boxsep=1pt, left=0pt,right=0pt,top=0pt,bottom=0pt,colframe=cyan!20, colback=cyan!20]{Duodenogastric reflux and bile reflux gastritis $\to$ Treatment $\to$ Calcium carbonate.} \tcbox[on line, boxsep=1pt, left=0pt,right=0pt,top=0pt,bottom=0pt,colframe=red!30, colback=red!30]{Stomach reflux $\to$ Related disease $\to$ Excess gastric acid.} \tcbox[on line, boxsep=1pt, left=0pt,right=0pt,top=0pt,bottom=0pt,colframe=cyan!20, colback=cyan!20]{Caved-S $\to$ Indication $\to$ Excess gastric acid.} \tcbox[on line, boxsep=1pt, left=0pt,right=0pt,top=0pt,bottom=0pt,colframe=cyan!20, colback=cyan!20]{Weile tablets $\to$ Indication $\to$ Excess gastric acid.}

         \vspace{5pt}
        
        {\textbf{\#\#\# Answer:}} \\
        \tcbox[on line, boxsep=1pt, left=0pt,right=0pt,top=0pt,bottom=0pt,colframe=red!30, colback=red!30]{Stomach reflux} can be a symptom caused by \tcbox[on line, boxsep=1pt, left=0pt,right=0pt,top=0pt,bottom=0pt,colframe=cyan!20, colback=cyan!20]{gastroesophageal reflux disease (GERD)}, and this condition could potentially lead to reflux esophagitis. Consider the following medical recommendations:
        \tcbox[on line, boxsep=1pt, left=0pt,right=0pt,top=0pt,bottom=0pt,colframe=cyan!20, colback=cyan!20]{Acid-suppressing medications (Proton Pump Inhibitors)}: You may consider using omeprazole or esomeprazole to effectively reduce gastric acid secretion, aiding in relieving symptoms of gastric reflux and mitigating the development of reflux esophagitis. Alternatively, you can use \tcbox[on line, boxsep=1pt,left=0pt,right=0pt,top=0pt,bottom=0pt,colframe=green!20, colback=green!20]{acid-neutralizing } \tcbox[on line, boxsep=1pt,left=0pt,right=0pt,top=0pt,bottom=0pt,colframe=green!20, colback=green!20]{medications (antacids) such as magnesium aluminum carbonate (Rennie), calcium carbonate, and aluminum hydroxide.} Another option is the use of \tcbox[on line, boxsep=1pt, left=0pt,right=0pt,top=0pt,bottom=0pt,colframe=cyan!20, colback=cyan!20]{H2 receptor} \tcbox[on line, boxsep=1pt, left=0pt,right=0pt,top=0pt,bottom=0pt,colframe=cyan!20, colback=cyan!20]{antagonists such as ranitidine or famotidine.} Additionally, medications like \tcbox[on line, boxsep=1pt, left=0pt,right=0pt,top=0pt,bottom=0pt,colframe=cyan!20, colback=cyan!20]{Caved-S or Weile tablets} are also suitable for mild to moderate symptoms of gastric reflux.
        Gastric protective medications: For short-term relief of stomach pain and discomfort, \tcbox[on line, boxsep=1pt, left=0pt,right=0pt,top=0pt,bottom=0pt,colframe=cyan!20, colback=cyan!20]{you may also consider using gastric mucosal protectants such as sucralfate.} It helps form a protective mucous layer, reducing irritation to the stomach mucosa.
        Diet and lifestyle adjustments: First, $\ldots$ it is advisable to consult a healthcare professional for evaluation and diagnosis as soon as possible.
        }
        \end{flushleft}
    \end{minipage}
};
\node[fancytitle, rounded corners, right=10pt] at (box.north west) {A case study of HyKGE pipeline};
\end{tikzpicture}%
\label{tab:case study}
\end{tabular}
\end{figure*}

\subsection{Ablation Study (RQ 2)}
\label{Ablation Study (RQ 2)}
To answer RQ2, we perform ablation studies to verify the effectiveness of the critical components of HyKGE, as illustrated in Table~\ref{tab:comparison}. Our observation can be summarized as follows:

\textbf{In pre-retrieval phase.} When we remove the Hypothesis Output Module, results are even deteriorating than base model. This is attributed to the fact that retrieved knowledge simply based on user queries is either insufficient or futile because of lacking direction for exploration. 
Nevertheless, the results of \texttt{w/o HO} are still better than KGRAG and we argue the reason is the reranking of reasoning chains effectively filters out noise during the post-retrieval phase. 

\textbf{In post-retrieval phase.} The removal of the Reranker leads to a noticeable decline in performance compared to HyKGE, which indicates that Reranker effectively eliminates excessive noise introduced by the retrieved knowledge, retaining only the most pertinent parts for answering the question. When we use entire $\mathcal{HO}$ and $\mathcal{Q}$ instead of chunk $(\mathcal{Q} \oplus \mathcal{HO})$ to perform reranking with reasoning chains, a decline in performance is also observed. This is attributable to the misalignment between dense retrieved knowledge and sparsely distributed keywords in $\mathcal{HO}$ and $\mathcal{Q}$, inducing a tendency to select more general or lengthier knowledge, thereby diminishing the \texttt{HOM}'s capability to supplement diverse knowledge.

Moreover, results of \texttt{w/o Chains} and \texttt{w/o Description} demonstrate that even when KG lacks certain knowledge, descriptive information or relevant knowledge chains can still enhance the answering capabilities of LLMs, which is believed to be associated with the inherent implicit knowledge within the LLMs themselves.

\subsection{Interpretability Analysis (RQ 3)}
\label{rq3}
In this section, we concentrate on evaluating the interpretability with three metrics \textbf{ACJ}, \textbf{BLEU}, \textbf{PPL} and \textbf{ROUGE-R} as shown in Table~\ref{tab:comparison2} to find out whether the retrieved knowledge is effective and whether it can help LLMs reduce hallucinations. Several observations can be derived from the results.

\textbf{The relevance of knowledge retrieval.}
For methods that interacted with LLMs and applied noise filtering modules, such as QE, CoK, CoN, SuRE and HyKGE, we notice that they often score higher on ACJ on MMCU-Medical and CMB-Exam, and ROUGE-R on CMB-Clin dataset, reflecting the efficacy of the LLMs' inherent knowledge and reasoning abilities as well as the importance of removing irrelevant knowledge. 
Moreover, the ACJ value of KG-GPT and QE is the second-to-best as they do not alter the semantics of the user query. Therefore, the knowledge retrieved by KG-GPT and QE have higher relevance with ACJ score, compared to CoK and CoN. Furthermore, it is noticed that our proposed HyKGE surpasses baselines with a performance gain of \textbf{84.19\%-378.71\%} and \textbf{133.33\%-345.22\%} on MMCU-Medical and CMB-Exam respectively, which demonstrates our superiority in solving the misaligned knowledge density between user query and retrieved knowledge. The marked decline in ACJ of \texttt{w/o Fragment} also supports the HO Fragment Granularity-aware reranker's role in keeping relevant knowledge. The BLEU and ROUGE-R scores on CMB-Clin also demonstrate HyKGE's superiority, indicating that HyKGE could be more appropriate for and aligned with real-life doctor consultations, proving the effectiveness of HyKGE in information retrieval.

\textbf{Can LLMs utilize retrieved knowledge to reduce hallucinations?}
As for method KGRAG, it fails to perform well on PPL and ROUGE-R, which is attributed to the provision of overly lengthy retrieved knowledge and redundant noise, resulting in the inability of the LLMs to extract useful information from the knowledge. The performance test of baselines consistently shows that our proposed HyKGE greatly reduces hallucinations and promotes LLMs to better utilize the retrieved knowledge, with performance gain of \textbf{57.77\%-95.36\%} and \textbf{61.03\%-185.42\%} on MMCU for PPL and ROUGE-R respectively. We argue the reason that the retrieved knowledge is more relevant and diverse because of the \texttt{HOM} and HO Fragment Granularity-aware Reranker, and its chain structure also stimulates the reasoning ability of LLMs. Other methods such as QE, CoN, and CoK's have been greatly reduced because their rerankers cannot retain more diverse knowledge, resulting in LLMs' answers being too singular and ROUGE-R surely being lower.
Notably, our performance on the CMB-Exam test set was superior, due to its richer and more detailed description of medical questions, allowing us to obtain more diverse and relevant knowledge based on $\mathcal{HO}$ and $\mathcal{Q}$.
\begin{figure}[htb]
  \centering
  % \vspace{-0.45cm}
  \includegraphics[scale=0.214]{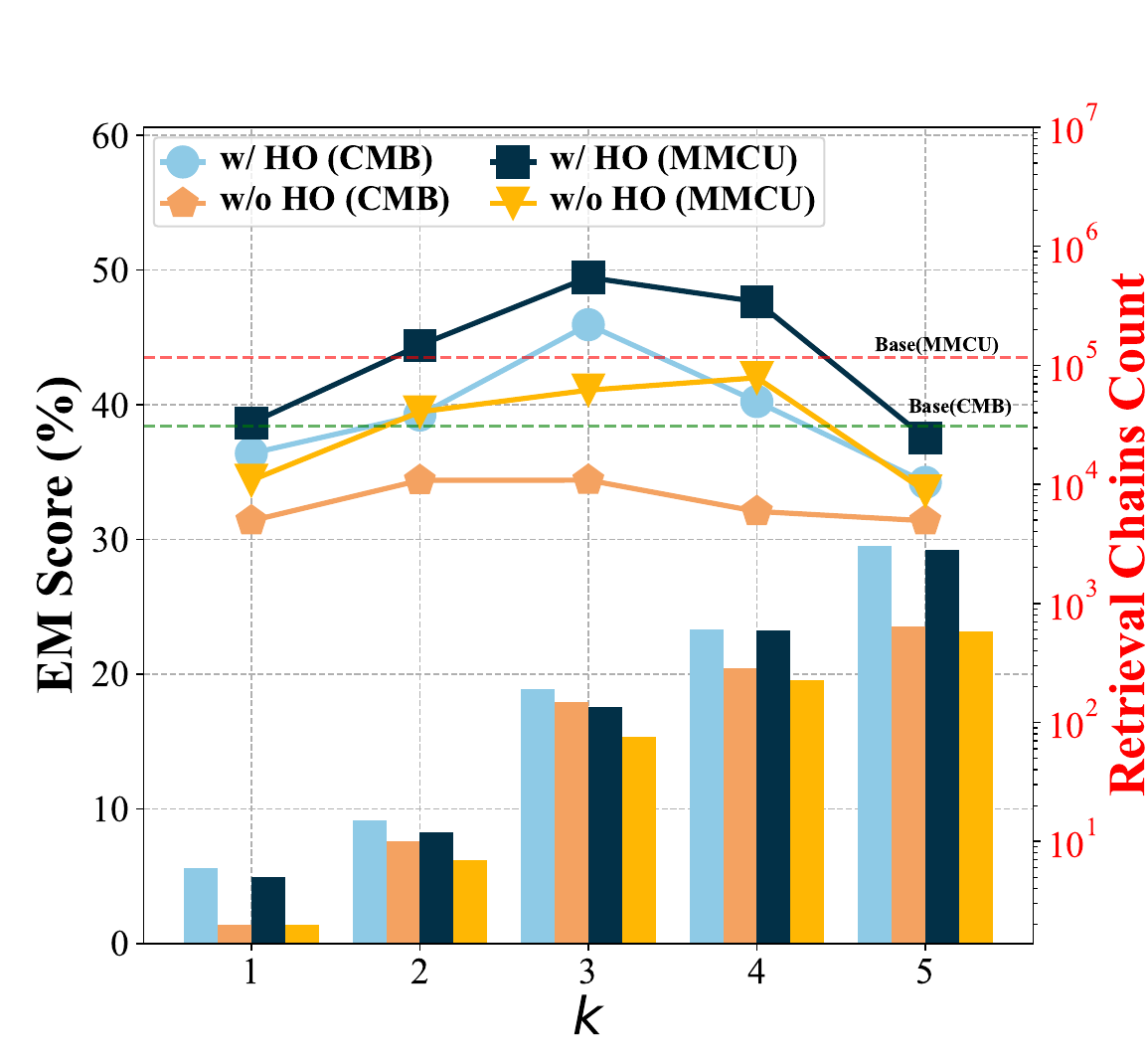}
  \includegraphics[scale=0.2314]{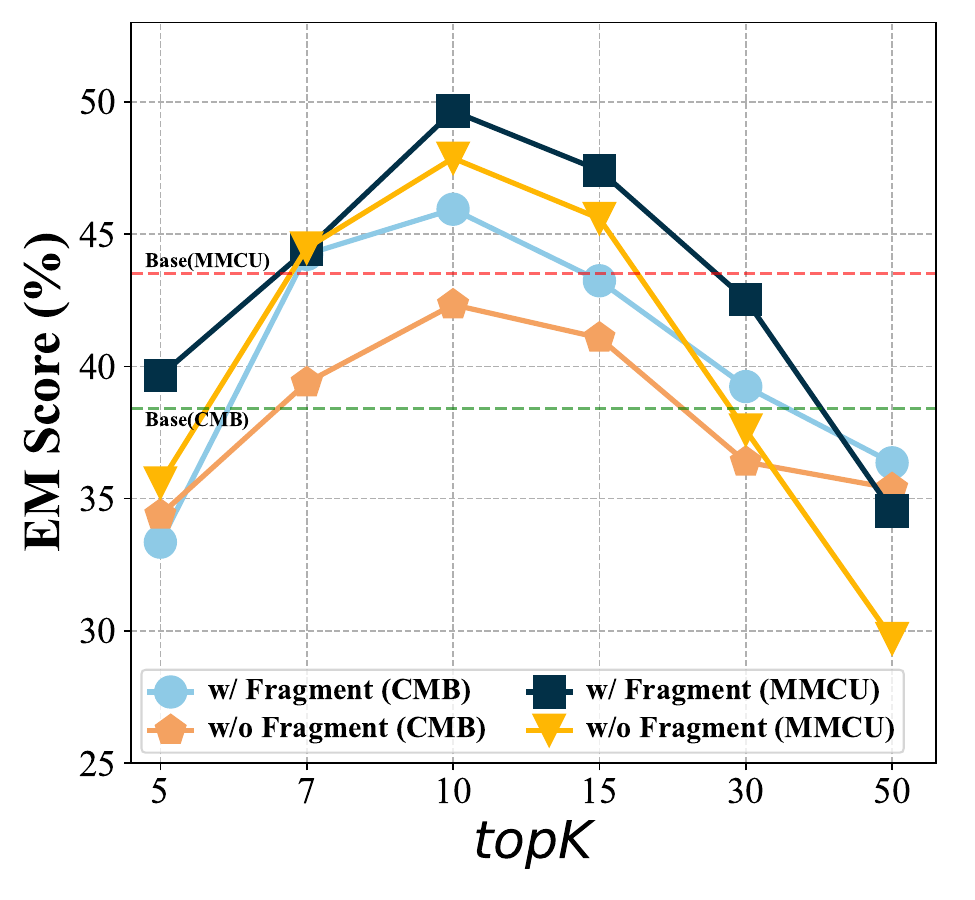}
  % \vspace{-5pt}
  \vspace{-0.25cm}
  \caption{(\textbf{Left}.) Hyper-parameter study with the KG hop $k$ on MMCU-Medical and CMB-Exam with GPT 3.5 turbo, from 1 to 5. (\textbf{Right}.) Hyper-parameter study with the reranker $topK$ on MMCU-Medical and CMB-Exam with GPT 3.5 turbo, from 5 to 50.
  }
   \label{fig:hyper_study}
   \vspace{-0.4cm}
   % \vspace{-10pt}
\end{figure}

\subsection{Hyper-parameter Study (RQ4)}
\label{rq4}
In this part, we concentrate on evaluating the influence of different hyper-parameters on HyKGE for RQ4. Specifically, we perform a series analysis of KG hop $k$ from the list $[1, 2, 3, 4, 5]$ and reranker $topK$ from the list $[5, 7, 10, 15, 30, 50]$ to verify the sensitive: 

Figure~\ref{fig:hyper_study} (Left.) depicts EM and the number of retrieved knowledge before pruning. We observe that as $k$ increases, the amount of knowledge retrieved explodes exponentially following a power-law distribution\cite{power_lay_2,power_lawer_1}, exceeding $10^3$ when $k=5$. However, an excessive amount of knowledge not only fails to improve EM, but also burdens LLMs with an increased number of tokens. Concurrently, EM exhibits a trend of initial increase followed by a decrease as $k$ increases. This phenomenon can be attributed to the fact that at lower values of $k$, the retrieved knowledge predominantly consists of isolated snippets of information, offering minimal utility. Conversely, with larger $k$, the LLMs encounter limitations in comprehending extensive reasoning chains, thereby rendering them incapable of effectively utilizing the complex and abundant retrieved information, with the performance even worse than the base model. Besides, we notice removing the HO will result in a substantial reduction in the quantity of knowledge retrieved, because of the limited diversity of knowledge obtained based solely on user query.

Figure~\ref{fig:hyper_study} (Right.) depicts EM with different reranking thresholds. Similar to Figure~\ref{fig:hyper_study} (Left.),  as $topK$ increases, the trends demonstrate that overwhelming reasoning chains will hamper LLMs' ability for comprehension. Meanwhile, it is obvious that HyKGE \texttt{w/o Fragment} always underperforms on EM as analyzed in Section~\ref{Ablation Study (RQ 2)}.

\subsection{Case Study (RQ2 and RQ3)}
\label{rq23}
This case study presents a representative sample that illustrates the effectiveness of our HyKGE model using GPT-3.5 Turbo as shown in Table~\ref{tab:case study}. The color coding within the table is key to understanding the source and validity of the information and we have these observations: 
\textbf{\underline{i)}} Compared to a brief user query, semantic spaces of $\mathcal{HO}$ are more abundant and have a clear direction for answering, helping us better understand user intention and extract more effective entity information. Ultimately, HyKGE extracted 23 entities from $\mathcal{HO}$ compared to only 1 from $\mathcal{Q}$. \textbf{\underline{ii)}} Comparing the $\mathcal{RC}$ with $\mathcal{RC}_{\texttt{prune}}$, it can be observed that the pre-filtered chains contain a large amount of irrelevant or repetitive knowledge, marked in black. After reranking, retrieved knowledge is highly non-redundant and relevant to $\mathcal{HO}$ and $\mathcal{Q}$, demonstrating the effectiveness of our fragment-based reranker. Ultimately, out of 125 reasoning chains, HyKGE selected $topK=10$ of the most valuable chains. \textbf{\underline{iii)}} Note that retrieved knowledge effectively assisted LLMs in correcting errors, mitigating the issue of hallucinations. In $\mathcal{HO}$, LLMs posited that ``\textit{calcium carbonate could not treat GERD}''; however, with the supplemental knowledge about ``\textit{calcium carbonate}'' in our retrieved reasoning chains, marked in green. LLMs corrected this error in its final response. In general, this case study demonstrates HyKGE's strong ability to generate hypotheses and validate them against a structured KG, effectively leveraging $\mathcal{HO}$ for exploring and reasoning chains for error correction. In general, the integration of these components ensures that the RAG's outputs are not only contextually relevant but also accurate, showcasing the interpretability and potential for AI-assisted decision-making in healthcare. 

\subsection{Efficiency Analysis (RQ2)}
\label{subsection:Component Efficiency Analysis}
To illustrate the effectiveness of our HyKGE module, we conducted a comparative analysis of the time overhead between HyKGE and other knowledge graph-enhanced LLM approaches, as presented in Table~\ref{table1}. The KGRAG method demonstrates the shortest time overhead among RAG methods, as it solely necessitates conveying the retrieved knowledge to the LLM Reader. However, when juxtaposed with QE and HyKGE, KGRAG's performance notably lags behind, even resulting in a negative gain because of the huge noise. In contrast to QE, HyKGE incurs slightly higher time primarily due to the noise filtering process, which consumes some time. Nonetheless, the performance enhancement achieved by HyKGE outweighs this marginal increase in time overhead.
Furthermore, CoN and CoK, which adopt the chain-of-thought strategy~\cite{wei2023chainofthought}, entail multiple interactions with LLMs, which proves to be considerably restrictive, particularly in real-world medical Q\&A scenarios where time is a critical consideration. Therefore, striking a balance between time overhead and model accuracy becomes imperative, in which regard HyKGE emerges as the most efficient and high-performing framework.

Moreover, inspired by these Chain-of-thought works~\cite {pouplin2024retrievalaugmented,trivedi2023interleaving}, which respectively employ LLMs in different processes of RAG, we embarked on similar endeavors. Specifically, we integrated LLMs into the modules of NER, Reranker, and summarization modules (to summarize the retrieved knowledge)~\cite{chen2024improving}, as shown in Table~\ref{tab:comparison3}. However, our findings underscored that leveraging such large-parameter models for tasks amenable to smaller counterparts incurs substantial time costs with marginal benefits. For instance, incorporating LLMs into NER, aimed at enhancing entity extraction, a task that could be efficiently handled by specialized pre-trained medical NER models, not only doubling interaction time but also introducing complexities such as misinterpretation of instructions, thus impeding subsequent processing. Similarly, the utilization of LLMs in Reranker considerably strained token resources. For instance, upon retrieving the query ``\textit{I feel stomach reflux after eating. What medicine should I take?}'' it generated a whopping 125 reasoning chains. However, employing LLMs to eliminate noisy knowledge from these chains resulted in decreased effectiveness. We argue that this was primarily due to the inundation of tokens, causing LLMs to lose in the middle~\cite{liu2023lost}, thereby impeding their ability to discern genuinely relevant knowledge from the retrieved chains and even ignore LLMs' tasks. Consequently, LLMs employed for Reranker inadvertently filtered out valuable knowledge, yielding negative outcomes and exacerbating computational overhead. Likewise, employing LLMs for knowledge summarization encountered challenges akin to those encountered in Reranker. 
Although LLMs are quite effective, according to Occam's razor principle~\cite{Standish_2004}, it is not always beneficial to use LLMs in every RAG step. Excessive reliance on LLMs can only lead to wasted time costs.
In summary, because RAG involves a process of continuous trial and error~\cite{barnett2024seven}, we experimented with many strategies and ultimately arrived at HyKGE.

\section{Conclusion}
In this paper, we proposed HyKGE, a hypothesis knowledge graph enhanced framework for LLMs to improve accuracy and reliability.  
In the pre-retrieval phase, we leverage the zero-shot capability of LLMs to compensate for the incompleteness of user queries by exploring searching directions through hypothesis outputs. In the post-retrieval phase, HyKGE applies a fragment reranking module to enhance the knowledge density alignment between user queries and retrieved knowledge, preserving relevant and diverse knowledge chains.
The comprehensive experiments conducted on three medical Q\&A tasks with two LLMs turbo demonstrate the effectiveness of HyKGE. 
Nevertheless, it remains worthwhile to contemplate how to dynamically optimize fragment granularity in the post-retrieval phase—a direction that we are committed to exploring actively in the future. In addition, despite the limitations of data sources and the high computational cost of LLMs, we will experiment on more other language or domain-specific KGs in the future to enhance the scalability and generalization of HyKGE.

% TODO: CMB-Clin数据集的分析+实验表格

\newpage
\balance
\bibliographystyle{ACM-Reference-Format}
\bibliography{acmart}

\end{CJK*}
\end{document}